\documentclass[journal]{IEEEtran}
\usepackage{graphicx}
\usepackage{amsmath}
\usepackage{amsthm}
\usepackage{amsfonts}
\usepackage{booktabs}
\usepackage{algorithm}
\usepackage{algorithmic}
\usepackage{subfigure}
\usepackage{multirow}
\usepackage{makecell}
\usepackage{fancyhdr}
\usepackage{pifont}
\usepackage{color}
\usepackage[colorlinks,linkcolor=black]{hyperref}
\UseRawInputEncoding

\begin{document}
%
\title{Relation-Based Associative Joint Location for Human Pose Estimation in Videos}
%
%
%

\author{Yonghao Dang, Jianqin Yin,~\IEEEmembership{Member,~IEEE}, and Shaojie Zhang
\thanks{Corresponding author: Jianqin Yin}
\thanks{Yonghao Dang, Jianqin Yin, and Shaojie Zhang are with the School of Artificial Intelligence, Beijing University of Posts and Telecommunications, Beijing 100876, China (e-mail:dyh2018@bupt.edu.cn; jqyin@bupt.edu.cn; zsj@bupt.edu.cn) }
}

%
%

\markboth{Journal of \LaTeX\ Class Files,~Vol.~14, No.~8, August~2015}%
{Shell \MakeLowercase{\textit{et al.}}: Bare Demo of IEEEtran.cls for IEEE Journals}

\maketitle

\begin{abstract}
  Video-based human pose estimation (VHPE) is a vital yet challenging task. While deep learning algorithms have made tremendous progress for the VHPE, lots of these approaches to this task implicitly model the long-range interaction between joints by expanding the receptive field of the convolution or designing a graph manually. Unlike prior methods, we design a lightweight and plug-and-play joint relation extractor (JRE) to explicitly and automatically model the associative relationship between joints. The JRE takes the pseudo heatmaps of joints as input and calculates their similarity. In this way, the JRE can flexibly learn the correlation between any two joints, allowing it to learn the rich spatial configuration of human poses. Furthermore, the JRE can infer invisible joints according to the correlation between joints, which is beneficial for locating occluded joints. Then, combined with temporal semantic continuity modeling, we propose a Relation-based Pose Semantics Transfer Network (RPSTN) for video-based human pose estimation. Specifically, to capture the temporal dynamics of poses, the pose semantic information of the current frame is transferred to the next with a joint relation guided pose semantics propagator (JRPSP). The JRPSP can transfer the pose semantic features from the non-occluded frame to the occluded frame. The proposed RPSTN achieves state-of-the-art or competitive results on the video-based Penn Action, Sub-JHMDB, PoseTrack2018, and HiEve datasets. Moreover, the proposed JRE improves the performance of backbones on the image-based COCO2017 dataset. Code is available at \href{https://github.com/YHDang/pose-estimation}{https://github.com/YHDang/pose-estimation}.
\end{abstract}

\begin{IEEEkeywords}
  Human pose estimation, keypoint detection, relation modeling, temporal consistency.
\end{IEEEkeywords}

%
\IEEEpeerreviewmaketitle

\section{Introduction}

 \IEEEPARstart{V}{ideo-based} human pose estimation (VHPE) plays an important role in computer vision. Results from the pose estimation can largely influence the subsequent high-level analysis of human action in videos, such as action recognition \cite{7961774}, action assessment \cite{7298734}, and human-computer interaction \cite{ComputerInteraction}. Therefore, it is essential to improve the performance of human pose estimation to facilitate vision tasks.

 Current human pose estimation methods mainly fall into two categories: regression methods and body part detection methods \cite{2019Deep}. The former aims at directly regressing coordinates of body joints from the image. The latter seeks to produce approximate coordinates of joints with the guidance of the heatmaps representation \cite{2019Deep}. It is challenging to regress the coordinate from the entire image. Therefore, heatmap regression of joints has been the prevalent method for human pose estimation due to its superior performance \cite{Luo2020}. Although human pose estimation (HPE) \cite{HRNet,9008554,HPE,SimpleBaseline,8100073,CSM,SPM,UDP,DarkPose,ViPNAS} has made a significant breakthrough, there is still a major factor that is important but neglected, i.e., most approaches model the relationship between joints implicitly by expanding the receptive field of convolutions. It is difficult for them to flexibly capture the complex and variable relationships among the joints.

 \begin{figure}[htbp]
   \centering
   \includegraphics[scale=0.3]{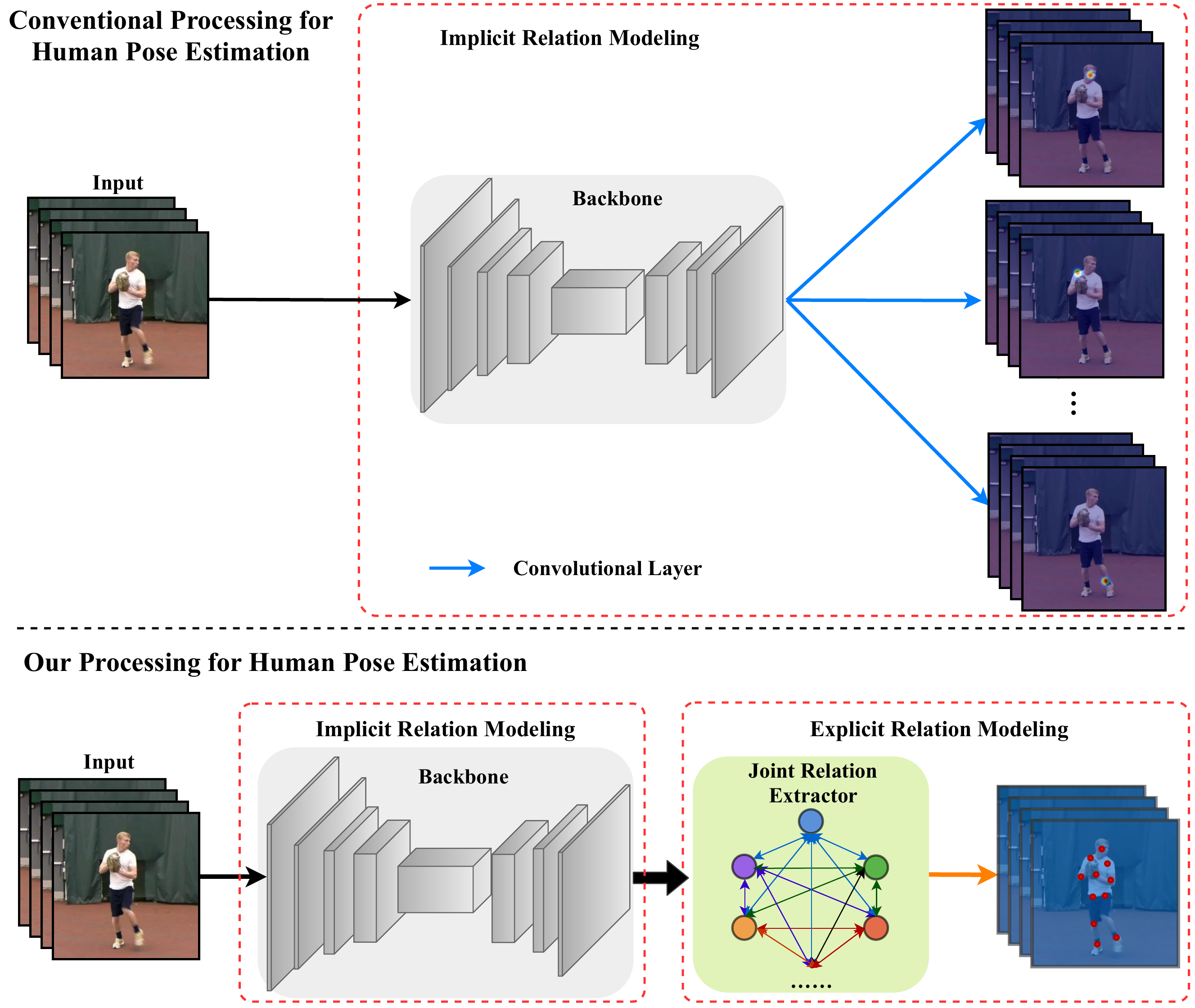}
   \caption{The difference between the proposed method and conventional methods. Upper: traditional HPE approaches expand the receptive field of convolutions to implicitly model the long-range interaction between joints. Lower: our proposed RPSTN uses an additional joint relation extractor to explicitly and automatically model the correlation between joints. Circles of varying colors represent distinct joints in the JRE module.}
   \label{fig:comparison}
 \end{figure}

 The structural information of poses is beneficial for locating body joints due to the topology of the human body. There are different correlations between any two body joints when people are in motion. These correlations reflect the structural information of different poses. In the case of occlusion, the model can infer occluded joints based on the correlation between joints. As a result, while estimating human poses, the correlation between joints should be taken into account. Although some works adopt the manual tree-based \cite{DLCM,8953713,8482293} or graph-based \cite{2020Graph,Repose} structure to model relationships between joints or parts. These methods rely on the hand-crafted structures. By using the prior of human bodies, these models can well learn the relationship defined on the prior structure. However, the predefined structures make it difficult to describe the complex relationships of the poses. In contrast to existing methods, we propose a lightweight and plug-and-play relation-based generator called a joint relation extractor (JRE) to explicitly and automatically model the correlation between any two joints, as shown in the lower of Fig. \ref{fig:comparison}. In this way, the proposed JRE module can infer the position of invisible joints according to the correlation between joints. 

 Besides, temporal semantic continuity is essential to estimate poses in videos. The pose semantic features should include the spatial configuration and the appearance features about poses. Because there is a minor difference between two adjacent frames, semantic features of poses in the previous frame can be employed to help locate invisible joints in the current frame. \cite{9008554} proposed a pose kernel distillator (PKD) to distill and transfer each joint's appearance features and positional information between adjacent frames. Although \cite{9008554} performs well, it ignores the explicit modeling and propagation of the relationship between joints. When people are moving, body joints intercat with one another. It is necessary to transfer the relation of joints from the current frame to the next frame. Inspired by \cite{9008554}, we design a joint relation guided pose semantics propagator (JRPSP) to assist the model in learning the temporal semantic continuity of poses. Different from \cite{9008554}, our method can explicitly propagate the additional association of joints due to the JRE module. Features propagated by JRPSP module contain appearance features, positions of joints, and the relationship between any two joints. We can regard the features extracted by the JRPSP module as the relation-based semantic features of the entire pose.

 In this paper, we present a Relation-based Pose Semantics Transfer Network (RPSTN) to automatically and explicitly model the associative relationship between joints for video-based human pose estimation. There are two critical components in the RPSTN: a joint relation extractor (JRE) to model spatial configurations of poses and a relation-based pose semantics propagator (JRPSP) to capture temporal dynamics from pose sequences. Specifically, the JRE module takes pseudo heatmaps (that include the spatial representation and positional information of joints) as input and models the correlation between any two joints by measuring the similarity of pseudo heatmaps. Besides, the JRPSP module extracts relation-based semantic features of the whole pose from the current frame and transfers these features to the next frame. The proposed RPSTN can infer invisible joints based on the structural information of poses learned by the JRE module. Furthermore, RPSTN can transfer the pose semantic features from the non-occluded frame to the occluded frame to locate occluded joints. The main contributions of our work are summarized as follows.
 \begin{itemize}
    \item We design a plug-and-play and lightweight relation-based generator (the number of parameters is $2 \times K \times K$, where $K$ is the number of joints) called joint relation extractor (JRE) to generate joint heatmaps by explicitly and automatically modeling the spatial correlation between any two joints. The JRE module is suitable for the mainstream video-based and single image-based pose estimation frameworks.
    \item We propose a Relation-based Pose Semantics Transfer Network (RPSTN) that can infer the position of the occluded joint by using the correlation between the occluded joint and its related joints in the spatial domain and transferring the pose information from the non-occluded frame to the occluded frame in the temporal domain.
    \item The proposed RPSTN achieves state-of-the-art or competitive results on four challenging video-based datasets, including the Penn Action, Sub-JHMDB, PoseTrack2018, and HiEve datasets. Besides, we verify the effectiveness of the proposed JRE module on the image-based COCO2017 dataset, which shows that JRE can improve the performance of different pose estimation backbones.
 \end{itemize}

\section{Related Works}

 \subsection{Relation modeling for pose estimation}

 Some works relied on hand-crafted features and pictorial structures to model the pose structural information. Lee et al. \cite{MCMC2004} applied a data-driven Markov chain Monte Carlo (MCMC) framework to search the solution space. This technique, however, was highly dependent on the input features. Andriluka et al. \cite{PSR2009} used an undirected graph to model the relationship among the connected joints. Yang and Ramanan \cite{APE-FMP2011} developed a novel mixture model for capturing contextual co-occurrence relations between body parts. To estimate human poses, the methods outlined above employed hand-crafted features that had to be constructed by the researcher. The subjective elements of researchers have a influence on the hand-crafted characteristics.


 With the development of convolutional neural networks (CNNs), many works have recently achieved superior performance for estimating human pose from static images \cite{HRNet,9008554,SimpleBaseline,UDP,DarkPose,ViPNAS,8482293,7961778,8578840,HigherHRNet,MIPNet,RLE,HiEve}. However, these methods implicitly model the interaction between joints by enlarging the receptive field of the convolution. To directly model the structural prior of poses, some researchers have combined the CNN with hierarchical tree-based structures to model the relation between body parts \cite{DLCM,8953713,8482293}. In these methods, joints were manually divided into several groups based on the distance between them. These approaches model joints' relationships inside the same group, but lack of modeling joint interactions between groups. Some works have been using manual graphs to model the relationship between joints. Wang et al. \cite{2020Graph} proposed Graph-PCNN which employs a graph designed manually to refine heatmaps and uses fully connected layers to regress the position of joints. Isack et al. \cite{Repose} proposed RePose with a manual graphic module to extract the relational information of body joints. In \cite{UnitPart2021}, two attitudes of the part-based and the holistic pose predictions were combined to estimate poses. Bin et al. \cite{PGCN2020} constructed a directed graph between body keypoints based on the natural compositional model of a human body and offered a local PGCN and a non-local PGCN to capture the local and long-rang dependencies of joints. Although the approaches discussed above perform well in terms of human pose estimation, they rely on the hand-crafted structures, which hinders them from learning complicated and changeable interactions between joints. In contrast, our proposed JRE module can explicitly and automatically model the correlation between any two joints, allowing the JRE to learn the rich structural information of poses spontaneously.

 \subsection{Temporal features modeling in human pose estimation}

 Modeling temporal information is significant for video-based human pose estimation. Because the optical flow can convey apparent velocities of movement \cite{TCE2020}, it is frequently used to describe the temporal features of videos \cite{8100073,2020Graph}. However, it is hard for the optical flow to deal with the large appearance variations due to the occlusion and motion blur. There were also methods applying 3D convolution to model the spatiotemporal representation of videos. For example, Girdhar et al. \cite{GirdharGTPT18} expanded the 2D Mask R-CNN to 3D for estimating the human pose from videos. Wang et al. \cite{CDT} used 3D HRNet and a spatial-temporal merging procedure to locate joints. To model the long-range temporal representation, LSTM \cite{1997LSTM} was used to model the dynamic information of poses. Luo et al. \cite{8578644} combined LSTM with the convolutional pose machines (CPM) \cite{7780880} to extract spatiotemporal features of poses. Artacho et al. \cite{Unipose2020} used the waterfall atrous spatial pooling (WASP) and LSTM to model sufficient contextual information. These methods aim to learn the temporal evolution of poses, while ignoring the temporal semantic continuity of videos. Actually, the pose information in the previous frame can be reused in the current frame to help locate joints due to the slight difference between two adjacent frames.

 Xu et al. \cite{ViPNAS} proposed ViPNAS to search networks and temporal feature fusions between adjacent frames for fast online video pose estimation. ViPNAS can search for temporal connections and choose which layer features to fuse automatically. It models the temporal semantic continuity indirectly by fusing the the features of continuous frames. Nie et al. \cite{9008554} formulated the video-based pose estimation as the feature matching procedure to transfer pose features between adjacent frames. Li et al. \cite{TCE2020} proposed a novel temporal consistency exploration to explicitly model videos' temporal consistency. However, features extracted or transferred by above methods lack the pose structural information that is beneficial for locating joints. In contrast to these methods, the proposed method explicitly learns the pose structural information by modeling correlations between joints and transfers the pose structural information between continuous time.

\section{Methodology}

 This section begins with an overview of the proposed RPSTN. The key components, joint relation extractor (JRE) and improved joint relation guided pose semantics propagator (JRPSP), are then described in detail.

\subsection{Relation-based Pose Semantics Transfer Network}

\begin{figure}[htbp]
    \centering
    \includegraphics[scale=0.42]{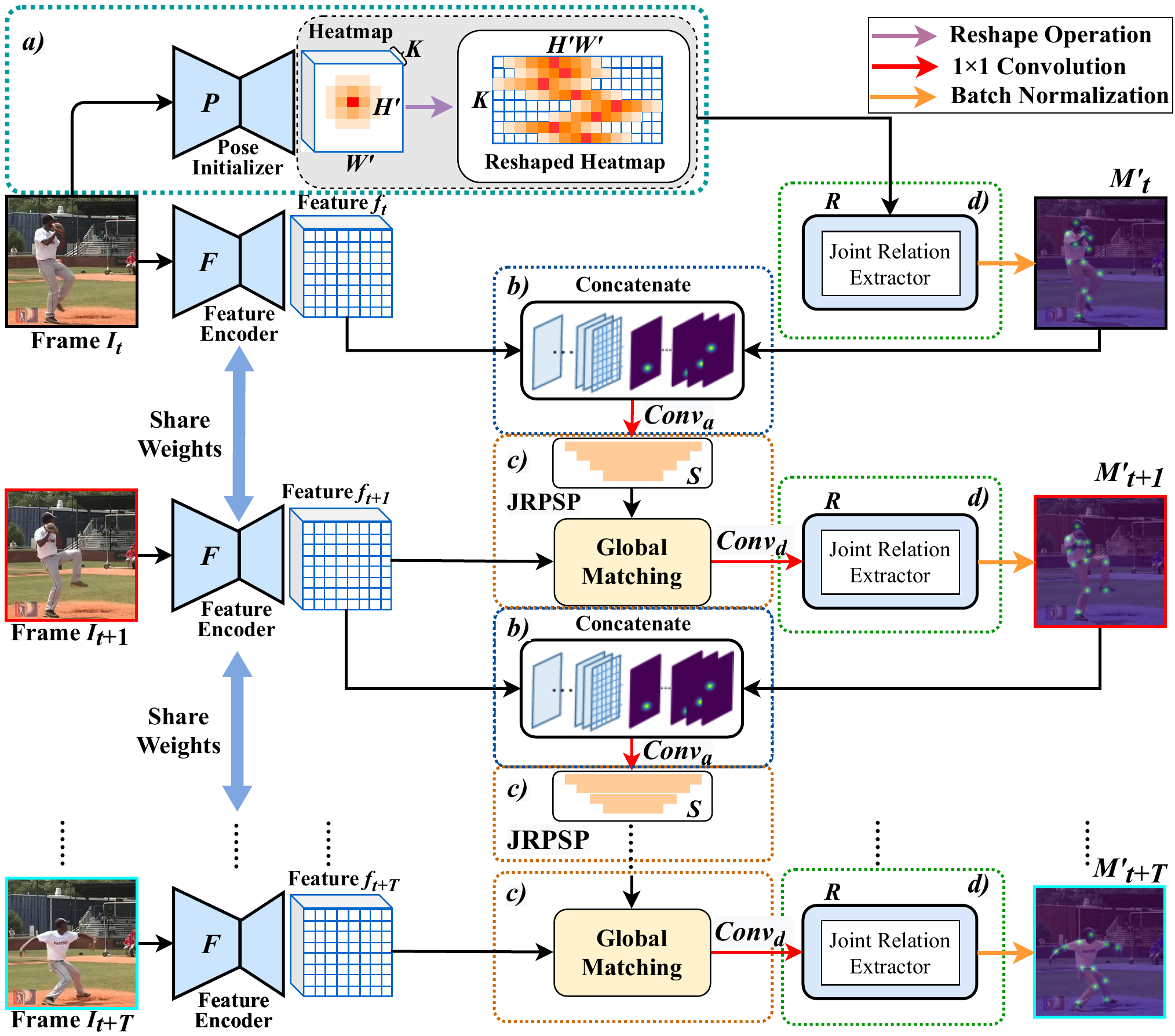}
    \caption{Overview of our proposed RPSTN. The pose initializer $P$ generates the initial pose for the first frame. The Joint Relation Extractor (JRE) adopts the initial heatmaps produced by $P$ as the input. It transfers initial heatmaps to accurate heatmaps by modeling the correlation between any two joints. For future frames, features and heatmaps in the previous frame are concatenated and passed to the JRPSP module to obtain high-level semantic features of poses. These features are then used to match similar regions in the current frame. Finally, the JRE module generates joint heatmaps by explicitly and automatically modeling the correlation between joints .}
    \label{fig:SPTNet}
 \end{figure}

 Different from \cite{9008554}, this paper proposes a joint relation extractor (JRE) to explicitly and automatically model the correlational relationship between joints, as well as a joint relation guided pose semantics propagator (JRPSP) designed for JRE to model the temporal semantic continuity of videos. Our framework can explicitly transfer the correlation between joints due to the JRE module. Furthermore, we follow the framework proposed in \cite{9008554}, as shown in Fig. \ref{fig:SPTNet}. Although except for the JRE and JRPSP, RPSTN uses the similar framework as \cite{9008554}, for clearly and completely introducing the proposed method, we describe the RPSTN from five aspects: a) initial pose estimation in the first frame, b) historical pose semantic learning in the previous frame, c) pose semantic propagation and global matching, d) the additional relation modeling in the current frame, and e) the loss function.

 Given a clip of the video that includes $N$ frames, i.e., $V=\left\{ {{I_t}} \right\}_{t = 1}^N,{I_t} \in {\mathbb{R}^{H \times W \times 3}}$, the RPSTN takes $T$ continuous frames as input, as shown in Fig. \ref{fig:SPTNet}.

 \paragraph{\textbf{Initial pose estimation}} For the first frame, a pre-trained pose initializer is used to produce the initial pose represented by a series of heatmaps. Then the JRE module takes initial heatmaps as input and explicitly learns the correlation between any two joints. The initial heatmaps are transformed into final heatmaps by the JRE module. We can formalize this process as follows.

 \begin{equation}
  {{M'}_t} = R(P({I_t})), t=1
  \label{eq:frame1}
 \end{equation}
 where ${M'}_t (t=1)$ is the heatmaps of joints in the first frame. $R(\cdot), P(\cdot)$ are JRE module and pose estimator.

 \paragraph{\textbf{Historical pose semantic learning}} To describe the processing of the proposed RPSTN, we only consider the $t$-th frame here. The discriminative semantic features of historical poses are useful for locating joints in the current frame. Pose semantic features should include the positional information as well as the appearance features about poses. To learn the historical semantic features of the entire pose, we concatenate feature maps $f_{t}$ and heatmaps $M'_{t}$ of the $t$-th frame along the channel and use a $1 \times 1 \times C$ convolution (where $C$ represents the number of channels) to aggregate the positional information and appearance features.

 \begin{equation}
 X_t^a = {Conv}_a(f_{t} \oplus {{M'}_t})
 \label{eq:conv1_1}
 \end{equation}
 where $X_t^a$ is the feature aggregated by ${Conv}_a$ in the $t$-th frame, ${Conv}_a$ represents $1\times1\times C$ convolution that is used to fuse the appearance features and positional information of joints, $\oplus$ denotes the concatenate operation. Features of the $t$-th frame ($f_{t}$) include appearance feature of joints. Heatmaps in the $t$-th frame (${M'}_t$) include the positional information of joints and the correlation between joints. Therefore, $X_t^a$ contains appearance features,  positional information of joints, and the correlation between joints. $X_t^a$ can be considered as the semantic features of poses.


 \paragraph{\textbf{Pose semantic propagation and global matching}} Due to the temporal semantic consistency of videos, the pose knowledge of the $t$-th frame can be reused to help locate joints in the $(t+1)$-th frame. The global matching mechanism is used to transfer the pose information between adjacent frames by searching regions in the $(t+1)$-th frame that have similar semantics with the $t$-th frame. In practice, the JRPSP takes the semantic features of poses $X_t^a$ as input to further distill relation-based semantic features of the whole pose in the $t$-th frame. JRPSP consists of several convolutional layers (that will be described in detail in subsection $C$). Therefore, with the expansion of receptive field, JRPSP can capture the long-range spatial dependency of poses. Features extracted by JRPSP $S(X_t^a)$ can be regarded as high-level semantic features of the entire pose.

 For the global matching, we take the high-level semantic features of the whole pose, $S(X_t^a)$ in the $t$-th frame, as templates and the features of the $(t+1)$-th frame, $F({I_{t + 1}})$, as targets. The template is used as the dynamic convolution \cite{DynamicConv} to make convolution with the target. Features $S(X_t^a)$ contain the positional information of joints in the $t$-th frame. After the dynamic convolution, the area around joint's position in the $(t+1)$-th frame will be excited. Furthermore, $F({I_{t + 1}})$ contains spatial representation of joints in the $(t+1)$-th frame. The other $1 \times 1 \times K$ convolution ${Conv}_d$ (where $K$ denotes the number of joints) is used to produce pseudo heatmaps of joints. Inheriting $S(X_t^a)$ and $F({I_{t + 1}})$, features $M_{t + 1}$ also contain the positional information of joints in $t$-th frame and the spatial representation of joints in $(t+1)$-th frame. Therefore, we regard features $M_{t + 1}$ as the pseudo heatmaps of joints, which contains rich semantic information of joints.

 \begin{equation}
  M_{t + 1} = {Conv}_d(S(X_t^a) \otimes F({I_{t + 1}}))
  \label{eq:conv1_2}
 \end{equation}
 where $M_{t+1}$ denotes pseudo heatmaps of joints in the $(t+1)$-th frame.  ${Conv}_d$ and $S(\cdot)$ represent $1\times1\times K$ convolution and the JRPSP module. $\otimes$ denotes the dynamic convolution.


 \paragraph{\textbf{Relation modeling in the current frame}} The JRE module is in charge of modeling the correlational relationship between joints in a frame. The JRE takes pseudo heatmaps of joints $M_{t + 1}$ as input and generates final joint heatmaps $M'$ by learning the associative relationship between joints. The JRE module is the main contribution of this paper, and its details will be introduced in subsection $B$.

 \begin{equation}
 {{M'}_{t + 1}} = BN(R(M_{t + 1}))
 \label{eq:heatmaps}
 \end{equation}

 \paragraph{\textbf{Loss function}} We adopt the generic Mean Square Error (MSE) loss to train the proposed model. The loss function is defined as

 \begin{equation}
   loss = \frac{1}{T}\sum\limits_{t = 1}^T {{l_2}({{M'}_t} - \widehat {{{M'}_t}})}
 \label{eq:Loss}
 \end{equation}
 where ${M'}_t$ is the heatmap generated by the model. $\widehat{{M'}_t}$ is the heatmap with the Gaussian distribution that is generated according to the ground truth.

\subsection{Joint Relation Extractor}

\begin{figure}[htbp]
    \centering
    \includegraphics[scale=0.5]{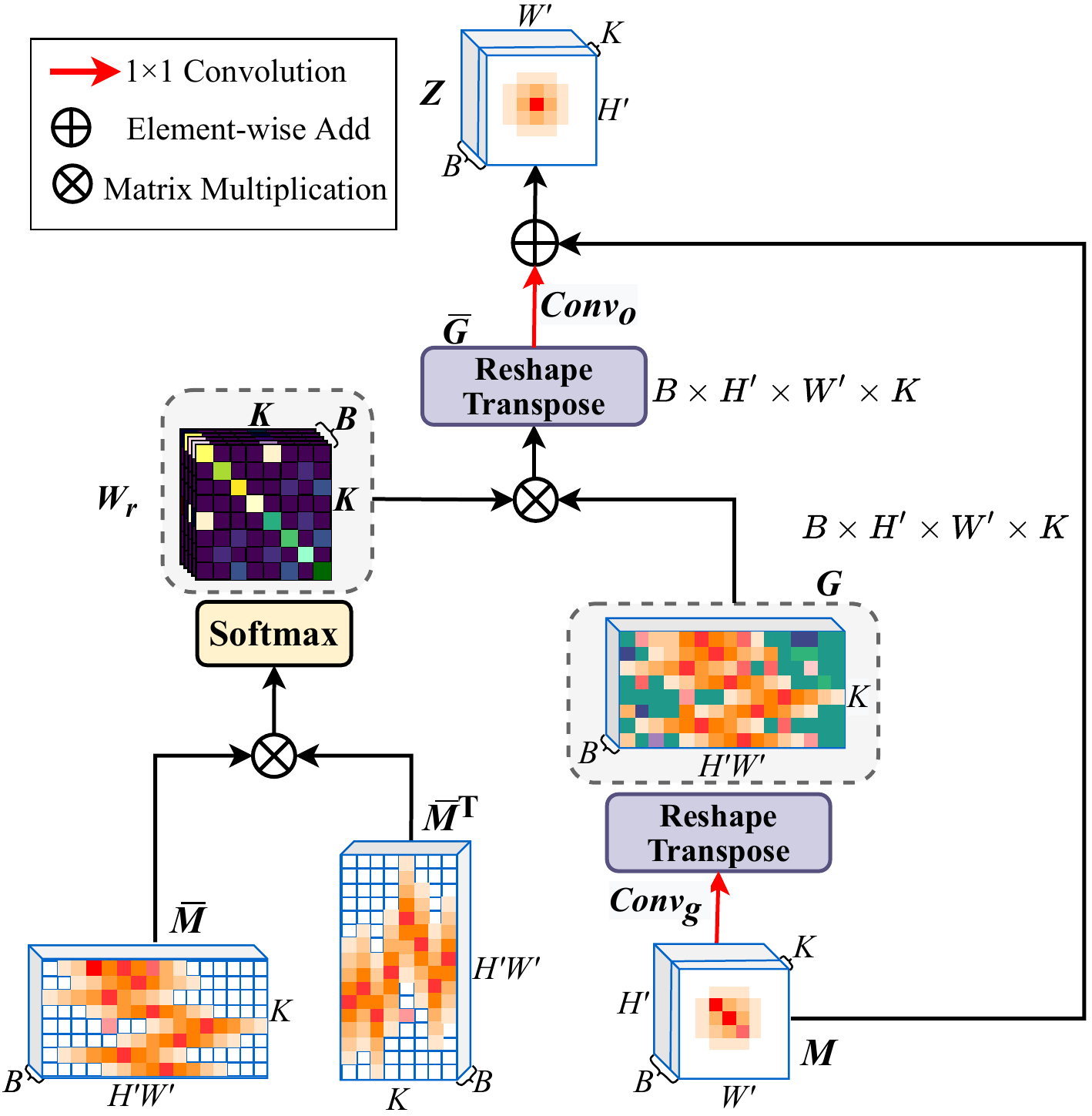}
    \caption{The structure of the joint relation extractor (JRE). $\otimes$ and $\oplus$ represent the matrix multiplication and the element-wise add. The JRE module takes the pseudo heatmap $M$ as the input. To model the correlation between body joints, $M$ is reshaped to $\bar{M}$, and a copy of $\bar{M}$ is transposed to $\bar{M}^T$. Each row in $\bar{M}$ or each column in $\bar{M}^T$ represents a joint. Then the multiplication of $\bar{M}$ and $\bar{M}^T$ is used to calculated the correlation between any two joints. The correlation matrix is applied to excite the structural information of the $G$ that aggregates the global features across all joints from the pseudo heatmap $M$.}
    \label{fig:REModule}
 \end{figure}

 \paragraph{\textbf{Description of the JRE module}} Correlations between joints constrain their positions due to the direct or indirect connection. The correlation between joints, which can be thought of as structural information of the pose, is useful for locating joints. To model the structural information of poses, we propose a lightweight and plug-and-play joint relation extractor (JRE) module that explicitly and automatically models the correlation between any two joints, as shown in Fig. \ref{fig:REModule}.

%

 \textbf{Joint correlation modeling.} As stated in subsection $A$, the pseudo heatmap $M$ inherits the spatial features $F({I_{t + 1}})$ and the high-level semantic features of the pose $S(X_t^a)$ which comprise spatial representation and positional information of joints. As a result, the pseudo heatmap $M$ also contains extensive joint semantic information, such as the spatial representation and positional information of the joint. To explicitly model the correlation between any two joints, each pseudo heatmap in $M$ is flattened and transposed to a row or column vector denoted as $\bar{M}$ or $\bar{M}^T$, respectively. Each row in $\bar{M}$ and each column in $\bar{M}^T$ represent a joint. The dot-product can represent the correlation between two vectors \cite{8578911}. Therefore, the relational weight matrix of joints can be calculated as follows.

 \begin{equation}
    W_{r} = Softmax\left( {{{\bar{M}}^T} \cdot \bar{M}} \right)
    \label{eq:att}
 \end{equation}
 where $ {{\bar{M} }^T} \cdot \bar{M}$ denotes the correlation matrix, with each element representing the correlation between any two joints. $W_{r}$ is the relational weight matrix of joints that models the correlation between the current joint and all joints. Softmax aggregates the correlation between the current joint and all joints. Therefore, $W_{r}$ contains the correlational relationship between the current joint and all joints. Based on the above correlation, $W_{r}$ can pay much attention to the joints that are related to the current joint. Thus, each element in $W_{r}$ represents the importance of the joint to locate current joint, i.e., attention weight.

 \textbf{Relation-based heatmap generation.} To learn the robust pose structural information, the association across all joints should also be considered. We use a $1 \times 1$ convolutional layer marked as $Conv_g$ to aggregate features across pseudo heatmaps of all joints. In this way, we can obtain different feature combinations about the whole pose.

 \begin{equation}
  G = Reshape(Trans({Conv}_g\left(M\right)))
 \end{equation}
 where $Trans$ is the transpose operation. $G$ represents the semantic features about the whole pose and contains each joint's positional information. Then, We use the relational weight $W_{r}$ to highlight the important joints with the high correlation in features $G$.


 \begin{equation}
  \bar{G} = Con{v_o}(W_r \cdot G)
  \label{eq:G}
 \end{equation}
 where $\bar{G}$ is the pose features activated by the relational weight matrix. Important regions in pose features $G$ are activated with the encouragement of the attention weight. The semantic information of joints is mixed after the activation. A $Conv_o$ is used to enhance the semantic information of the joint by aggregating the joint's features in $\bar{G}$.

 Following joint correlation modeling, the location information of joints is degraded. To keep the positional information of the joint, the residual connection is utilized after $Conv_o$ to introduce the original information. To summarize, the JRE module can be formulated as follows.

 \begin{equation}
  \begin{array}{l}
    Z = {Conv}_o\left( W_{r} \cdot {Conv}_g\left( M \right) \right) + M
    \end{array}
    \label{eq:REM}
 \end{equation}
 where $W_{r}$ and $Con{v_o}$ are used to stimulate the important correlated joins and learn the semantic information of poses, respectively. The closely related joints are activated by combining the original information with the features extracted by $Con{v_o}$.

 \begin{figure}[htbp]
    \centering
    \includegraphics[scale=0.4]{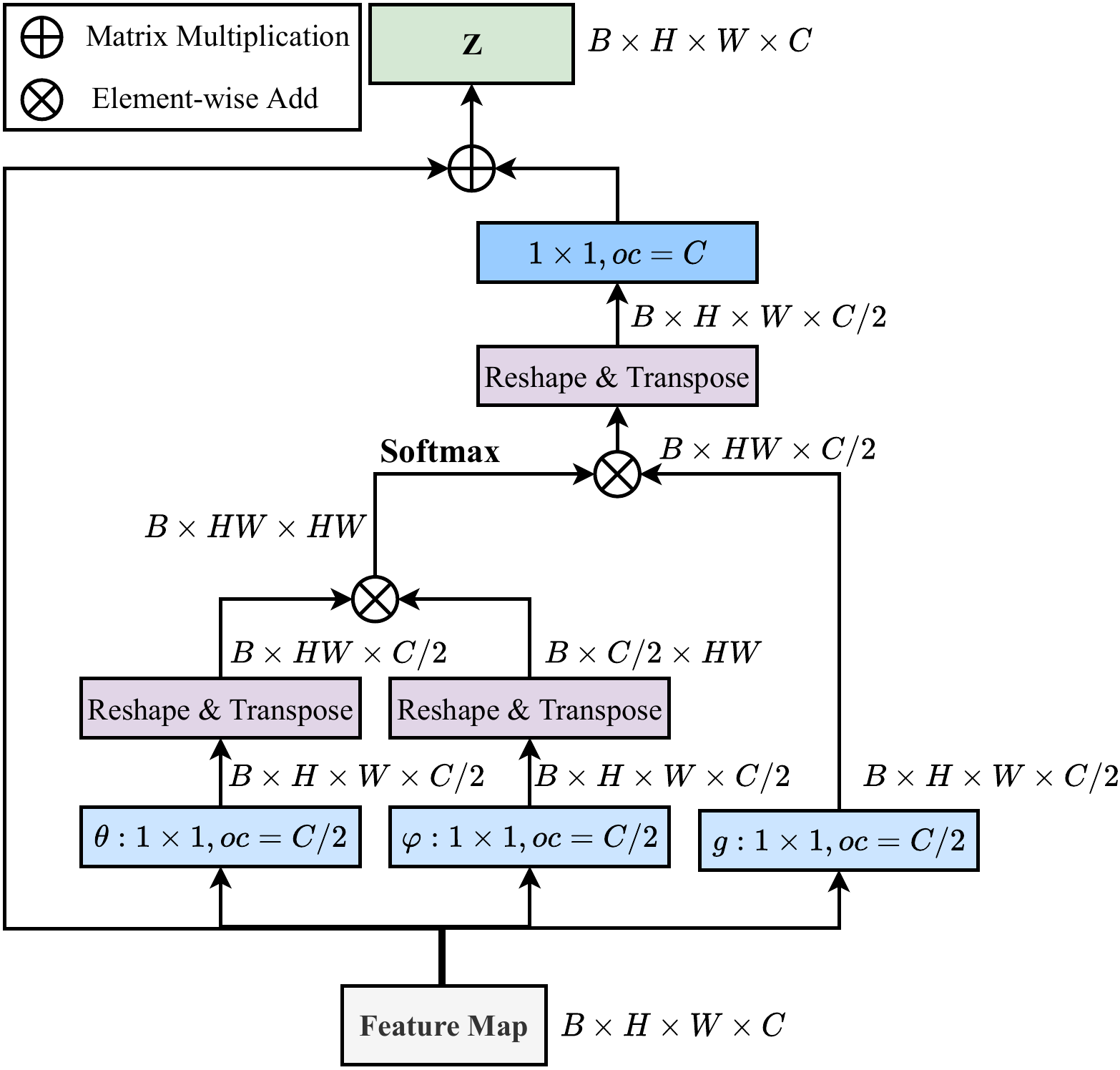}
    \caption{The structure of the original non-local block.}
    \label{fig:non-local}
 \end{figure}

 \paragraph{\textbf{Comparisons with Non-local Block}} We show the original non-local block in Fig. \ref{fig:non-local} to highlight differences between the non-local block and the proposed JRE module. Compared Fig. \ref{fig:REModule} with Fig. \ref{fig:non-local}, there are intrinsic differences between these two blocks. (1) Non-local block captures the dependency between pixels, while the proposed JRE models the correlation between pseudo heatmaps representing joints. That is to say, the non-local block models the pixel-level dependency, while the proposed JRE models the joint-level correlation. Specifically, as shown in Fig. \ref{fig:non-local}, the non-local block takes feature maps as input and produces the relational map with the size of $[B, HW, HW]$, where each element represents the relationship between two pixels. However, the JRE module takes the pseudo heatmaps as input and produces the relational map with the size of $[B, K, K]$ where each element represents the correlation between two joints. (2) The non-local block is used as an intermediate feature extractor to extract the global dependency of pixels. On the contrary, the JRE module aims at explicitly modeling the relationship between joints. Features used by JRE should contain the semantic information of each joint. Due to the rich semantic features in the deep layer of the network, the JRE module is usually used to help the model generate joint heatmaps. (3) The number of parameters of our JRE is lower than that of the non-local block. To reduce the computational cost, the input of the non-local block is copied to three groups and reduced dimension by three $1 \times 1$ convolutional layers, as shown in Fig. \ref{fig:non-local}. The pseudo heatmaps that contain rich semantic information of joints are enough to represent joints. Therefore, it is not necessary for JRE to extract further features or reduce computational costs by convolutions.


 \subsection{Joint Relation Guided Pose Semantics Propagator}

 Motivated by the DKD \cite{9008554}, we adopt a joint relation guided pose semantics propagator (JRPSP) to extract relation-based semantic features of the whole pose from the current frame and propagate these features to the next frame. Effectively transferring the semantic information of the whole pose between adjacent frames is vital to improve pose estimation. To model the semantic features of the entire pose, the propagator should have a larger receptive field and a deeper level to better represent the increased knowledge (structure) in the enlarged area (from joints to pose).

 \begin{figure}[htbp]
   \centering
   \includegraphics[scale=0.5]{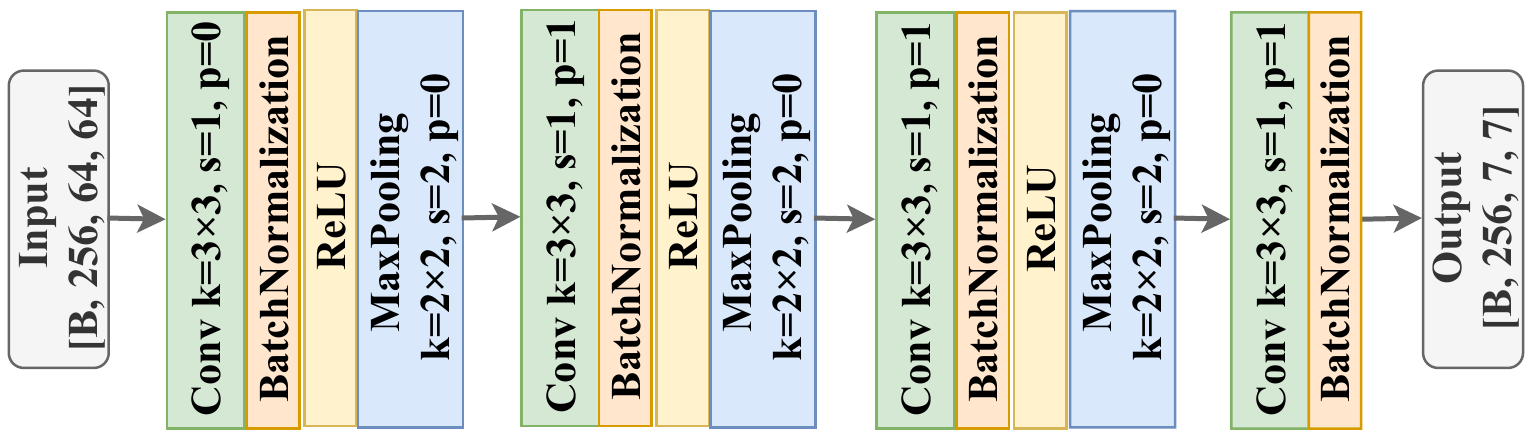}
   \caption{The structure of the joint relation guided pose semantics propagator (JRPSP).}
   \label{fig:SFEModule}
 \end{figure}

 Based on the DKD \cite{9008554}, we append a convolutional layer to enlarge the receptive field of the network, which allows the JRPSP to cover the enlarged area with the convolutional filters. As shown in Fig. \ref{fig:SFEModule}, JRPSP includes four convolutional layers and three pooling layers. The input of the JRPSP also includes appearance features, positional information, and the additional relational information of joints. Compared with the DKD, the input of the JRPSP contains more complex information. It is necessary to use a deeper network to extract effective features. With the gradual increase of the receptive field, the JRPSP can extract effective features.

\section{Experiments}

 \textbf{Datasets.} We evaluate the proposed RPSTN and the JRE module on video-based and image-based datasets, respectively.

 For the video-based dataset, we use the Penn Action \cite{6751390}, Sub-JHMDB \cite{JHMDB}, PoseTrack2018 \cite{PoseTrack18}, and HiEve \cite{HiEve} datasets. (1) \textbf{Penn Action dataset} contains 2326 video clips divided into two groups. Consistent with \cite{9008554}, one group includes 1258 clips for training, and the other contains 1068 clips for testing. Besides, annotations provide bounding boxes for people, position, and visibility for 13 body joints in each frame. (2) \textbf{Sub-JHMDB dataset} contains 316 video clips. Consistent with \cite{9008554}, we take 75\% clips for training and 25\% clips for testing. The Sub-JHMDB provides the complete bodies of people that contain 15 body joints. Different from the Penn Action, each joint in the Sub-JHMDB is annotated as visible. (3) \textbf{PoseTrack2018 dataset} is a large-scale dataset for multi-person human pose estimation. It contains 593 training samples, 170 validation samples, and 375 testing samples. The dataset provides 15 joints. (4) \textbf{HiEve dataset} is a large-scale and challenging human-centric dataset that is designed to help understand human behaviors in crowded and complex events. It has 32 video sequences containing 19 training videos and 13 testing videos. The dataset provides annotations for human poses in each frame and totally includes 1,099,357 human poses. There are 14 joints in each human pose.

 For the image-based dataset, we use the COCO2017 dataset \cite{COCO} to evaluate the proposed joint relation extractor. \textbf{COCO2017 dataset} consists of single images. It contains over $57K$ training images and $150K$ training person instances. Each person instance has 17 joints. We train the model on the COCO train2017 dataset and evaluate the model on the test-dev2017 dataset.

 \textbf{Implementation details.} For a fair comparison on the Penn Action and Sub-JHMDB datasets, we use the same setting as \cite{9008554}: the pose estimator and the feature extractor are implemented with SimpleBaselines (SBL) \cite{SimpleBaseline} and pre-trained on the MPII \cite{6909866} dataset; each training sample includes 5 continuous frames sampled from a video; the shape of each input frame is set to $256\times256$; the data augmentation is implemented following \cite{9008554}; the learning rate is initialized to 0.0005. In addition, we use Adam \cite{Adam} as an optimizer. The learning rate decreases by 0.1 at the 8th, 15th, 25th, 40th, and 80th epochs. Besides, we use the same experimental results as \cite{DCPose} on the PoseTrack2018 dataset.

 On the HiEve dataset, we use the Faster-RCNN \cite{Faster-RCNN} equipped with the ResNet-50 \cite{7780459} as the human detector. The Faster-RCNN is pretrained on the COCO2017 dataset and fine-turned on the HiEve training set. The SBL \cite{SimpleBaseline} with ResNet-50 is performed as the feature encoder of RPSTN. Besides, the SBL is pretrained on the COCO2017 and MPII datasets, and then fine-turned on the HiEve training set. To save the best model, we further split the training data into 15 training videos and 4 valid videos (including $3^{rd}, 7^{th}, 8^{th}, and 17^{th}$ videos) following \cite{ACMMM-HiEve}.

 On the COCO2017 dataset, we evaluate the proposed JRE on the mainstream pose estimation frameworks including SBL \cite{SimpleBaseline}, HRNet \cite{HRNet}, ViPNAS \cite{ViPNAS}, PoseFix \cite{PoseFix}, UDP\cite{UDP}, and DarkPose\cite{DarkPose}. Because PoseFix, UDP, and DarkPose are model-agnostic approaches, we use the same SBL with ResNet-50 as the pose initializer. For the raw PoseFix, it uses the output of SBL as the input. Here, we use the output of SBL with JRE as the input of PoseFix for comparison. Experimental settings are followed \cite{HRNet,SimpleBaseline,ViPNAS,UDP,DarkPose,PoseFix}, respectively. All experiments are conducted on two Nvidia GeForce GTX 1080Ti GPUs.

 \textbf{Evaluation.} For the video-based dataset, (1) the Penn Action and the Sub-JHMDB datasets are single-person datasets. Following \cite{9008554}, we adopt the commonly used Percentage of Correct Keypoints (PCK) \cite{PCK} as our metric (i.e., the position of a joint is considered to be correct if it falls within $\gamma \cdot L$ pixels of the ground truth \cite{9008554}). The $L$ is the reference distance and $\gamma$ is set to 0.2 to control the relative threshold \cite{8578644}. Following \cite{9008554}, the $L$ is set to $L = \max ({H_{box}},{W_{box}})$, where $H_{box}$  and $W_{box}$ are height and width of the person bounding box, and the size of torso which is represented by the Euclidean Distance between the left shoulder and the right hip \cite{8578644}, respectively. (2) PoseTrack2018 is a large-scale multi-person pose estimation and tracking dataset. It includes challenging situations in the crowded sense. Following \cite{ViPNAS,DCPose}, we evaluate the model for visible joints with the average precision (AP) metric that is used to evaluate multi-person human pose estimation. (3) Following \cite{HiEve}, we adopt average precision(AP@$\alpha$) and weighted average precision (w-AP@$\alpha$) as evaluation metrics on the HiEve dataset, where $\alpha$ represents the specific IOU threshold to determine true/false positive \cite{HiEve}. w-AP is proposed to assign larger weights to the image during evaluation if it has a higher crowd index and anomalous behavior to avoid the methods only focusing on simple cases or uncrowded scenarios in the dataset \cite{HiEve}.

 For the image-based COCO2017 dataset, following \cite{HRNet,SimpleBaseline,ViPNAS,UDP,DarkPose}, we also use the AP evaluation metric that is based on Object Keypoint Similarity (OKS) to evaluate the model. Different from the AP used on the PoseTrack2018, the APs used on COCO2017 are divided into five categories according to the target size and the range of the IOU.

\subsection{Ablation Studies}

 In this section, we first evaluate how the capacity of the proposed JRE module affects the performance of the model. Then we verify the effectiveness of the JRPSP. Next, we verify the influence of different backbones and the pose initialize. Finally, we show the influence of the number of frames used in temporal modeling. Experiments are conducted on the Penn Action dataset.

 \begin{table}[htbp]
   \footnotesize
   \setlength\tabcolsep{2.4pt}

   \caption{Ablation studies for JRE and JRPSP on the Penn Action dataset. The PCK normalized by torso size is used to be as the evaluation metric.}
   \begin{tabular}{lcccccccc}
   \hline
   Methods                  & Head & Sho. & Elb. & Wri. & Hip & Knee & Ank. & mPCK \\
   \hline
                            &\multicolumn{8}{c}{Ablation Studies for the JRE module} \\
   RPSTN-w/o-JRE-JRPSP      & 97.2 & 93.0 & 90.1 & 87.5 & 92.9 & 92.6 & 89.2 & 91.4 \\
   RPSTN-w/o-JRE            & 97.6 & 96.6 & 91.6 & 89.5 & 96.5 & 94.8 & 91.7 & 93.1 \\
   RPSTN-NL                 & 97.7 & 92.5 & 85.4 & 81.7 & 92.9 & 92.5 & 86.5 & 89.3 \\
   RPSTN-RePose             & 97.1 & 88.1 & 87.7 & 85.9 & 90.1 & 92.0 & 88.7 & 89.5 \\
   RPSTN-RF-Head            & 96.4 & 94.4 & \textbf{94.2} & 92.3 & 96.4 & 96.2 & 91.2 & 94.2 \\
   RPSTN-JRE-2Convs         & 96.8 & 94.3 & 92.4 & 90.0 & 94.7 & 93.9 & 86.5 & 92.4 \\
   \hline
                            &\multicolumn{8}{c}{Ablation Studies for the JRPSP module}\\
   RPSTN-w/o-JRPSP          & 97.6 & 96.8 & 92.6 & 90.1 & 96.4 & 95.9 & 94.0 & 94.2 \\
   RPSTN-DKD                & 97.1 & 96.1 & 92.5 & 91.1 & 96.2 & 94.6 & 86.3 & 93.1 \\
   RPSTN-JRPSP-5Convs        & 86.9 & 93.4 & 92.3 & 88.3 & 94.1 & 94.0 & 82.2 & 91.2 \\
   \textbf{RPSTN (ResNet18)}    & \textbf{97.8} & \textbf{96.7} & 94.0 & \textbf{92.5} & \textbf{96.5} & \textbf{95.2} & \textbf{95.2} & \textbf{95.4} \\
   \hline
   \end{tabular}

   \label{tab:Abliation1}
 \end{table}

 \begin{table}[htbp]
   \footnotesize
   \setlength\tabcolsep{3pt}
   \centering
   \caption{Ablation studies for the Feature Extractor and Pose Initializer on the Penn Action dataset. The PCK normalized by torso size is used to be as the evaluation metric.}
   \begin{tabular}{lcccccccc}
   \hline
   Methods                  & Head & Sho. & Elb. & Wri. & Hip & Knee & Ank. & mPCK \\
   \hline
   RPSTN-Res18-Cat          & 97.5 & 96.4 & 92.9 & 91.8 & 94.8 & 94.7 & 93.4 & 94.3 \\
   RPSTN-Res18-w/o-P        & 97.0 & 96.0 & 93.2 & 91.2 & 94.9 & 92.8 & 80.7 & 91.9 \\
   \textbf{RPSTN (ResNet18)}& \textbf{97.8} & \textbf{96.7} & \textbf{94.0} & \textbf{92.5} & \textbf{96.6} & \textbf{96.5} & \textbf{95.2} & \textbf{95.4} \\
   \hline
   RPSTN-Res50-Cat          & 97.9 & 97.2 & 94.8 & 92.1 & 96.8 & 96.0 & 93.5 & 95.3 \\
   RPSTN-Res50-w/o-P        & 98.1 & 96.6 & 94.6 & 92.6 & 96.1 & 96.3 & 91.9 & 94.9 \\
   \textbf{RPSTN (ResNet50)}& \textbf{98.2} & \textbf{96.9} & \textbf{95.2} & \textbf{93.2} & \textbf{96.6} & \textbf{95.7} & \textbf{95.0} & \textbf{95.7} \\
   \hline
   \end{tabular}

   \label{tab:Abliation2}
 \end{table}

 \begin{table}[htbp]
   \footnotesize
   \setlength\tabcolsep{2pt}
   \centering
   \caption{Ablation studies for different backbones on the Penn Action dataset. The PCK normalized by torso size is used to be as the evaluation metric.}
   \begin{tabular}{lcccccccc}
   \hline
   Methods                  & Head & Sho. & Elb. & Wri. & Hip & Knee & Ank. & mPCK \\
   \hline
   RPSTN-Res18-w/o-JRE      & 97.6 & 96.6 & 91.6 & 89.5 & 96.5 & 94.8 & 91.7 & 93.1 \\
   \textbf{RPSTN-Res18}& \textbf{97.8} & \textbf{96.7} & \textbf{94.0} & \textbf{92.5} & \textbf{96.6} & \textbf{96.5} & \textbf{95.2} & \textbf{95.4} \\
   \hline
   RPSTN-Res50-w/o-JRE      & 97.4 & 95.8 & 93.4 & 92.0 & 96.3 & 94.6 & 91.2 & 94.2 \\
   \textbf{RPSTN-Res50}& \textbf{98.2} & \textbf{96.9} & \textbf{95.2} & \textbf{93.2} & \textbf{96.6} & \textbf{95.7} & \textbf{95.0} & \textbf{95.7} \\
   \hline
   RPSTN-Res101-w/o-JRE     & \textbf{97.8} & 96.7 & 95.1 & 92.5 & 96.3 & \textbf{97.1} & 94.6 & 95.6 \\
   \textbf{RPSTN-Res101}    & \textbf{97.8} & \textbf{97.1} & \textbf{96.1} & \textbf{95.1} & \textbf{96.7} & \textbf{97.1} & \textbf{96.1} & \textbf{96.5} \\
   \hline
   RPSTN-HG-w/o-JRE         & \textbf{97.7} & 96.7 & 92.4 & 92.3 & \textbf{96.8} & 96.7 & 94.4 & 95.0 \\
   \textbf{RPSTN-HG}        & \textbf{97.7} & \textbf{97.3} & \textbf{92.5} & \textbf{94.7} & \textbf{96.8} & \textbf{96.8} & \textbf{95.8} & \textbf{95.9} \\
   \hline
   RPSTN-HRNet-w/o-JRE      & 98.0 & 95.4 & 96.7 & 96.1 & 97.2 & 96.7 & 96.6 & 96.6 \\
   \textbf{RPSTN-HRNet}     & \textbf{98.2} & \textbf{96.6} & \textbf{97.1} & \textbf{96.2} & \textbf{97.2} & \textbf{97.2} & \textbf{97.1} & \textbf{97.0} \\
   \hline
   \end{tabular}
   \label{tab:backbones}
 \end{table}

 \paragraph{\textbf{Ablation studies for JRE module}} 
 First, we demonstrate the effectiveness of JRE module. We remove the JRE module and denote it as RPSTN-w/o-JRE. Compared to the RPSTN-w/o-JRE, the mean PCK of RPSTN (ResNet18) is increased by 2.3\%, indicating that the correlation between joints is effective for estimating human poses. Furthermore, to verify the effectiveness of the JRE module separately, we remove the JRPSP and keep the JRE based on RPSTN as RPSTN-w/o-JRPSP, and remove the JRPSP and JRE as RPSTN-w/o-JRE-JRPSP. RPSTN-w/o-JRPSP outperforms RPSTN-w/o-JRE-JRPSP by 2.8\%, which shows that modeling correlations between joints is beneficial for estimating poses. Moreover, we compare the JRE with a refinement head with 2 convolutional layers (denoted as RPSTN-RF-Head) to verify the effectiveness of the JRE. As can be seen from TABLE \ref{tab:Abliation1}, RPSTN (ResNet18) outperforms RPSTN-RF-Head by 1.2\% mPCK. The possible reason is that the JRE explicitly models the correlation between any two joints, while the refinement head only implicitly learns the interaction between joints.

 Second, We replace the JRE module with the non-local block (i.e., RPSTN-NL) to demonstrate the applicability of the relation between joints for the human pose estimation. RPSTN (ResNet18) improves in all body joints, especially the wrist, with a boost of 10.8\%. The non-local implicitly models the relationship of joints by calculating the correlation between pixels in feature maps. Experimental results confirm that the joint-level correlation is more important than the pixel-level correlation for pose estimation.

 Third, to verify the effectiveness of learning the relational information automatically, we introduce the manual graphic structure proposed in \cite{Repose} and describe it as RPSTN-RePose. Compared with the RPSTN-RePose, our RPSTN(ResNet18) achieves an overall improvement of 5.9\%. Furthermore, our method can accurately locate joints with high flexibility, such as the wrist, elbow, and ankle joints. The possible reason is that correlations between joints are complicated. The manual structure only models the relationship between connected joints, which prevents the model from learning rich structural information of poses.

 Finally, to evaluate the effectiveness of the pseudo heatmaps, we use two additional convolutions to further extract features from the pseudo heatmaps and describe it as RPSTN-JRE-2Convs. RPSTN (ResNet18) outperforms RPSTN-JRE-2Convs by 3.0\% mPCK. Furthermore, the PCK of RPSTN (ResNet18) in locating each joint is higher than that of RPSTN-JRE-2Convs. The pseudo heatmaps contain the rich semantic information of the joint. In other words, we can recognize each joint from pseudo heatmaps. The semantic information is mixed after adding two convolutions, which is not conductive to modeling the relationship between joints.

 \paragraph{\textbf{Ablation studies for JRPSP module}} 
 First, we verify the effectiveness of the JRPSP. We remove the JRPSP module and denote it as RPSTN-w/o-JRPSP. Our RPSTN (ResNet18) has an overall improvement of 1.2\%, demonstrating that the pose's historical relation-based semantic features improve the pose estimation.

 Second, we replace the JRPSP with the pose kernel distillatory proposed in DKD \cite{9008554} (denoted as RPSTN-DKD) to illustrate the effectiveness of relation-based semantic features for pose estimation. By comparing RPSTN (ResNet18) with RPSTN-DKD, we can observe our RPSTN (ResNet18) achieves an overall increment of 2.3\% (95.4\% vs. 93.1\%). The possible reason is that the receptive field of the JRPSP module is larger than that of DKD so that the JRPSP can capture the long-range interactions between parts, which helps the JRPSP extract more discriminative semantic features of the whole pose.

 Finally, we evaluate how the receptive field of JRPSP affects the performance of the model. We append a convolutional layer to the JRPSP and denote it as RPSTN-JRPSP-5Convs. The receptive field of the JRPSP is enlarged to 54 (the size of the input feature maps is $64 \times 64$). The performance is decreased by 4.2\%. We conjecture that RPSTN-JRPSP-5Convs is confused by redundant background features, causing errors during locating joints in subsequent frames.


 \begin{table*}[htbp]
  \footnotesize
  \centering
  \caption{Ablation studies for the number of frames on the Penn Action dataset. The PCK normalized by torso size is used to be as the evaluation metric.}
  \begin{tabular}{l|c|c|cccccccc}
   \hline
   Methods  & Frames & FLOPs (G) &  Head &  Sho. &  Elb. &  Wri. &  Hip &  Knee &  Ank. &  mPCK \\
   \hline
   \multirow{3}*{RPSTN}  &
   5-frames & \textbf{89.62} & \textbf{98.2} & \textbf{96.9} &  95.2 &  93.2 &  96.6 &  95.7 &  95.0 &  95.7 \\
            &
   7-frames & 120.20 & 97.9 & 96.3 & 95.5 & \textbf{93.5} & \textbf{97.1} & \textbf{97.3} & 96.0 & 96.1 \\
            &
   10-frames & 166.07 & 98.0 & \textbf{96.9} & \textbf{95.6} & \textbf{93.5} & 96.9 & 97.0 & \textbf{96.3} & \textbf{96.2} \\
   \hline
   \multirow{2}*{RPSTN-w/o-JRE} &
   7-frames & 120.36 & 98.1 & 96.7 & 93.3 & 91.5 & 96.5 & 96.5 & 94.6 & 95.1 \\
            &
   10-frames & 166.30 & 97.6 & 97.3 & 94.1 & 92.5 & 96.4 & 96.2 & 93.4 & 95.2\\
   \hline
   \end{tabular}
   \label{tab:longterm}
 \end{table*}

\begin{table*}[htbp]
  \footnotesize
  \centering
  \caption{Comparisons with the state-of-the-art methods on the Penn Action dataset.}
  \begin{tabular}{lccc|cccccccc}
   \hline
   Methods  & Params (M) & FLOPs (G) & Time (ms) &  Head &  Sho. &  Elb. &  Wri. &  Hip &  Knee &  Ank. &  mPCK \\
   \hline
      &  &  &  &  \multicolumn{8}{c}{Normalized by person size} \\
   Park et al. \cite{6126552} &
   - & - & - & 62.8 &  52.0 &  32.3 &  23.3 &  53.3 &  50.2 &  43.0 &  45.3 \\
   Nie et al. \cite{7298734} &
   - & - & - & 64.2 &  55.4 &  33.8 &  24.4 &  56.4 &  54.1 &  48.0 &  48.0 \\
   Iqal et al. \cite{7961774} &
   - & - & - & 89.1 &  86.4 &  73.9 &  73.0 &  85.3 &  79.9 &  80.3 &  81.1 \\
   Gkioxari et al. \cite{Gkioxari} &
   - & - & - & 95.6 &  93.8 &  90.4 &  90.7 &  91.8 &  90.8 &  91.5 &  91.8 \\
   Song et al. \cite{8100073} &
   - & - & - & 98.0 &  97.3 &  95.1 &  94.7 &  97.1 &  97.1 &  96.9 &  96.5 \\
   Zhang et al. \cite{KPN2020} &
   - & - & - & 98.7 &  \textbf{98.7} &  97.0 &  95.3 &  \textbf{98.8} &  \textbf{98.7} &  98.6 &  98.0 \\

   Li et al. \cite{TCE2020} &
   - & - & - & \textbf{99.3} &  98.5 &  97.6 &  97.2 &  98.6 &  98.1 &  97.4 &  98.0 \\

   Luo et al. \cite{8578644} &
   231.5 & 70.98 & 25 & 98.9 &  98.6 &  96.6 &  96.6 &  98.2 &  98.2 &  97.5 &  97.7 \\
   DKD(ResNet50) \cite{9008554} &
   219.92 & 89.51 & 11 & 98.8 &  \textbf{98.7} &  96.8 &  97.0 &  98.2 &  98.1 &  97.2 &  97.8 \\

   \textbf{RPSTN (ResNet18)}   &
   133.37 &
   68.29 &
   9    &
   99.1 &
   \textbf{98.7} &
   98.6 &
   98.4 &
   \textbf{98.8} &
   \textbf{98.7} &
   \textbf{98.8} &
   \textbf{98.7} \\
   \textbf{RPSTN (ResNet50)}   &
   222.17 &
   89.62 &
   12    &
   99.0 &
   \textbf{98.7} &
   \textbf{98.8} &
   \textbf{98.5} &
   \textbf{98.8} &
   \textbf{98.7} &
   \textbf{98.8} &
   \textbf{98.7} \\
   \hline
      &  &  &  &  \multicolumn{8}{c}{Normalized by torso size} \\

   Luo et al. \cite{8578644} &
   231.5 & 70.98 & 25 & 96.0 &  93.6 &  92.4 &  91.1 &  88.3 &  94.2 &  93.5 &  92.6 \\

   DKD(ResNet18) \cite{9008554} &
   131.12 & 68.18 & 8 &95.7 &  90.0 &  92.2 &  89.4 &  86.8 &  92.3 &  89.5 &  90.6 \\

   DKD(ResNet50) \cite{9008554} &
   219.92 & 89.51 & 11 & 96.6 &  93.7 &  92.9 &  91.2 &  88.8 &  94.3 &  93.7 &  92.9 \\

   \textbf{RPSTN (ResNet18)} &
   133.37 &
   68.29 &
   9    &
   97.8 &
   96.7 &
   94.0 &
   92.5 &
   \textbf{96.6} &
   \textbf{96.5} &
   \textbf{95.2} &
   \textbf{95.4} \\

   \textbf{RPSTN (ResNet50)} &
   222.17 &
   89.62 &
   12    &
   \textbf{98.2} &
   \textbf{96.9} &
   \textbf{95.2} &
   \textbf{93.2} &
   \textbf{96.6} &
   95.7 &
   95.0 &
   \textbf{95.7} \\
   \hline
   \end{tabular}

   \label{tab:SOTA}
 \end{table*}


 \paragraph{\textbf{Ablation studies for pose initializer}} Results on the first frame are significant for estimating poses from subsequent frames. As shown in TABLE \ref{tab:Abliation2}, we conduct a series of experiments to verify whether the pose initializer is necessary. First, we connect the feature extractor and the JRE module by two $1 \times 1$ convolutional layers, which is denoted as RPSTN-ResX-Cat ('X' represents 18 or 50). The RPSTN-ResX-Cat is inferior to the RPSTN (ResNetX). The possible reason is that the RPSTN-ResX-Cat introduces the background information. Experimental results illustrate that the pure spatial positional information of joints contributes to modeling the relation between joints for JRE module.

 Second, we remove the pose initializer based on the RPSTN-ResX-Cat to manifest the effect of the pose initializer and term it as RPSTN-ResX-w/o-P. As shown in TABLE \ref{tab:Abliation2}, the performance of RPSTN-ResX-w/o-P is worse than that of RPSTN (ResNetX). Especially, the average PCK of RPSTN-Res18-w/o-P decreases a lot (91.9\% vs. 95.4\%). However, the mean PCK of RPSTN-Res50-w/o-P is less than 1\% (0.8\%) lower than that of RPSTN (ResNet50). The reason is that the RPSTN (ResNet18) needs the pose initializer to provide the pure positional information of joints to model the relation between joints due to the limited ability to extract pose appearance features. In contrast, the ResNet-50 can extract discriminative features from the human pose, so the RPSTN-Res50-w/o-P achieves comparable results with the RPSTN (ResNet50), even if without the pose initializer.

 \paragraph{\textbf{Ablation studies for different backbones}} We use three representative human pose estimation models, including SBL \cite{SimpleBaseline} with varying CNN architectures, Hourglass \cite{Hourglass}, and HRNet \cite{HRNet}, to demonstrate the compatibility of JRE module. Experimental results listed in TABLE \ref{tab:backbones} show that JRE provides significant performance gain to the models. This suggests a generic usefulness of the proposed JRE module.

 \paragraph{\textbf{Ablation studies for the number of frames}} In order to explore the influence of frames on the RPSTN, we use 7, 10 frames for long-term temporal modeling, respectively. Results are recorded in TABLE \ref{tab:longterm}.

 It can be seen from TABLE \ref{tab:longterm} that increasing the number of frames improves the performance of the RPSTN. However, the improvement of the performance is slight, while the computational cost has almost doubled. To further verify the significance of JRE, we also remove JRE and use more frames to model the temporal information, described as RPSTN-w/o-JRE. In the case of 10 frames, compared with RPSTN-w/o-JRE, the JRE brings 1\% improvement of mPCK (96.2\% vs. 95.2\%). It is noteworthy that RPSTN with 5 frames outperforms RPSTN-w/o-JRE with 10 frames. Experimental results prove the significance of the proposed JRE.

\subsection{Comparisons with State-of-the-arts}

 \paragraph{\textbf{Quantitative analysis on the Penn Action dataset}} For the PCK normalized by person size, it can be seen from TABLE \ref{tab:SOTA} that RPSTN achieves state-of-the-art results on each body joint. Compared with the DKD (ResNet50), the performance of RPSTN (ResNet18) is increased, a relative improvement of 0.9\% (98.7\% vs. 97.8\%), and the accuracy of locating each joint is improved. It is noteworthy that the DKD (ResNet50) \cite{9008554} uses ResNet-50 as the feature extractor, while our RPSTN (ResNet18) uses ResNet-18 to extract features. Similarly, our RPSTN (ResNet50) also outperforms the DKD (ResNet50), especially for the elbow and the wrist joints. Experimental results indicate that the correlational relationship between joints is beneficial for locating body joints. However, the RPSTN (ResNet18) achieves the same mPCK as the RPSTN (ResNet50). We think that human poses in the Penn Action dataset are pure and straightforward, and it is enough to employ a shallow network (ResNet-18) to extract distinct appearance features. But for joints with frequent motion (e.g., elbow and wrist), the deep network can extract more discriminative features that are helpful for detecting the joint.

 For the PCK normalized by torso size, we can see that our RPSTN also outperforms existing methods. From TABLE \ref{tab:SOTA}, we can observe that RPSTN (ResNet50) has 1.01\% parameters and 0.12\% FLOPs more than DKD (ResNet50), but the performance is improved by 2.8\%. Besides, both RPSTN (ResNet18) and DKD (ResNet18) apply ResNet-18 to extract features, while our model outperforms the DKD (ResNet18) by 4.8\% (95.4\% vs. 90.6\%). For the hip joint, compared with the DKD (ResNet50), the RPSTN achieves an improvement of 7.8\%. We think that the hip joint is easily occluded, such as turning the body or squatting. Because our RPSTN can infer the occluded joint according to the joints related to the occluded joint, the occluded hip can be inferred accurately with the help of other joints.

\begin{table}[htbp]
  \footnotesize
  \setlength\tabcolsep{3pt}
  \centering
  \caption{Comparisons with The State-of-the-art Methods on the Sub-JHMDB Dataset.}
  \begin{tabular}{lcccccccc}
  \hline
  Methods  & Head &  Sho. &  Elb. &  Wri. &  Hip &  Knee &  Ank. &  mPCK \\
  \hline
           &  \multicolumn{8}{c}{Normalized by person size} \\
  Park et al. \cite{6126552} &
  79.0 &  60.3 &  28.7 &  16.0 &  74.8 &  59.2 &  49.3 &  52.5 \\
  Nie et al. \cite{7298734} &
  80.3 &  63.5 &  32.5 &  21.6 &  76.3 &  62.7 &  53.1 &  55.7 \\
  Iqal et al. \cite{7961774} &
  90.3 &  76.9 &  59.3 &  55.0 &  85.9 &  76.4 &  73.0 &  73.8 \\

  Song et al. \cite{8100073} &
  97.1 &  95.7 &  87.5 &  81.6 &  98.0 &  92.7 &  89.8 &  92.1 \\

  Li et al. \cite{TCE2020} &
  \textbf{99.3} &  98.9 &  96.5 &  92.5 &  98.9 &  97.0 &  93.7 &  96.5 \\

  Zhang et al. \cite{KPN2020} &
  95.1 &  96.4 &  95.3 &  91.3 &  96.3 &  95.6 &  92.6 &  94.7 \\

  Luo et al. \cite{8578644} &
  98.2 &  96.5 &  89.6 &  86.0 &  98.7 &  95.6 &  90.9 &  93.6 \\

  DKD(ResNet50) \cite{9008554} &
  98.3 &  96.6 &  90.4 &  87.1 &  \textbf{99.1} &  96.0 &  92.9 &  94.0 \\
\textbf{RPSTN (ResNet50)}   &
  98.9 &
  \textbf{99.1} &
  \textbf{99.0} &
  \textbf{97.9} &
  97.8 &
  \textbf{97.8} &
  \textbf{97.3} &
  \textbf{97.4} \\
\hline
   &  \multicolumn{8}{c}{Normalized by torso size} \\
  Luo et al. \cite{8578644} &
  92.7 &  75.6 &  66.8 &  64.8 &  78.0 &  73.1 &  73.3 &  73.6 \\
  DKD(ResNet50) \cite{9008554} &
  \textbf{94.4} &  78.9 &  69.8 &  67.6 &  81.8 &  79.0 &  78.8 &  77.4 \\
\textbf{RPSTN (ResNet50)} &
  91.0 &
  \textbf{87.1} &
  \textbf{82.1} &
  \textbf{80.5} &
  \textbf{88.8} &
  \textbf{85.9} &
  \textbf{83.8} &
  \textbf{85.8} \\
\hline
\end{tabular}

\label{tab:hmdb}
\end{table}

 \paragraph{\textbf{Quantitative analysis on the Sub-JHMDB dataset}} TABLE \ref{tab:hmdb} shows comparisons of our RPSTN model with state-of-the-arts on the Sub-JHMDB dataset. For the PCK normalized by person size, we can observe that our RPSTN (ResNet50) achieves the new state-of-the-art performance with a mean PCK of 97.4\%. The PCK of RPSTN (ResNet50) for each joint is better than DKD, indicating that explicitly modeling and transferring the correlation between joints is beneficial for human pose estimation. For the PCK normalized by torso size, compared with DKD (ResNet50), our RPSTN (ResNet50) achieves about 8.4\% improvement of the average PCK (85.8\% vs. 77.4\%). The accuracy of joints except for the head joint is promoted. Our method provides accurate location for the flexible joint, such as elbow, wrist, and ankle joints. Appearance features of these joints are easy to be lost due to the occlusion. Our model can infer these flexible joints according to the joint related to them.

 \paragraph{\textbf{Quantitative analysis on the PoseTrack2018 dataset}} Experimental results are shown in TABLE \ref{tab:posetrack}. TABLE \ref{tab:posetrack} shows that RPSTN achieves the best mean accuracy. Especially for the head, RPSTN outperforms DCPose by 0.6\% AP. The RPSTN can accurately locate the position of the head by modeling the correlation between the head and other joints. Furthermore, the head changes slightly between adjacent frames, so features in the previous frame help locate the head. However, RPSTN achieves the same AP as DCPose for the wrist. The possible reason is that the wrist is prone to motion blur, which makes it challenging to extract spatial features. The correlation between the wrist and other joint points will be weakened.

 \begin{table}[htbp]
  \footnotesize
  \setlength\tabcolsep{3pt}
  \centering
  \caption{Comparisons with The State-of-the-art Methods on the PoseTrack2018 Dataset.}
  \begin{tabular}{lcccccccc}
  \hline
  Methods  & Head &  Sho. &  Elb. &  Wri. &  Hip &  Knee &  Ank. &  Mean \\
  \hline
  AlphaPose \cite{AlphaPose} &
  63.9 &  78.7 &  77.4 &  71.0 &  73.7 &  73.0 &  69.7 &  71.9 \\
  MDPN\cite{MDPN} &
  75.4 &  81.2 &  79.0 &  74.1 &  72.4 &  73.0 &  69.9 &  75.5 \\
  DCPose\cite{DCPose} &
  83.5 &  86.3 &  81.2 &  \textbf{75.6}&  77.5 &  \textbf{77.9} &  71.9 &  79.5 \\
\textbf{RPSTN (Ours)} &
  \textbf{84.1} &
  \textbf{86.4} &
  \textbf{81.4} &
  \textbf{75.6} &
  \textbf{77.6} &
  \textbf{77.9} &
  \textbf{72.0} &
  \textbf{79.6} \\
\hline
\end{tabular}

\label{tab:posetrack}
\end{table}

\begin{table*}[htbp]
  \centering
  \caption{Comparisons with The State-of-the-art Methods on the COCO2017 test-dev Dataset. The input size of all backbones is set to $256 \times 192$. The values in brackets represent the gain of the backbone after integrating JRE.}
  \begin{tabular}{l|c|c|llllll}
  \hline
  Methods  & Params & GFLOPs & AP &  $AP^{50}$ &  $AP^{75}$ &  $AP^{M}$ &  $AP^{L}$ &  AR \\
  \hline
   SBL \cite{SimpleBaseline} & 32.424M & 8.911 &
   70.0 &  \textbf{90.9} &  77.9 &  66.8 &  75.8 &  75.6 \\
  \textbf{SBL-JRE} & 32.425M & 8.912 &
   \textbf{70.2 ($\uparrow$ 0.2)} &  \textbf{90.9} &  \textbf{78.5 ($\uparrow$ 0.6)} &  \textbf{67.1 ($\uparrow$ 0.3)} &  \textbf{75.9 ($\uparrow$ 0.1)} &  \textbf{75.9 ($\uparrow$ 0.3)} \\
\hline
  HRNet-W32 \cite{HRNet} & 27.214M & 7.120 &
    73.5 &  92.2 &  \textbf{81.9} & 70.2 &  79.2 &  79.0 \\
  \textbf{HRNet-W32-JRE} & 27.215M & 7.121 &
   \textbf{73.6 ($\uparrow$ 0.1)} &  \textbf{92.3 ($\uparrow$ 0.1)} &  \textbf{81.9} & \textbf{70.3 ($\uparrow$ 0.1)} &  \textbf{79.4 ($\uparrow$ 0.2)} &  \textbf{79.0} \\
\hline
   S-ViPNAS \cite{ViPNAS} & 6.949M & 1.388 &
   70.3 &  90.7 &  78.8 & 67.3 &  75.5 &  \textbf{77.3} \\
  \textbf{S-ViPNAS-JRE}  & 6.950M & 1.389 &
   \textbf{70.5 ($\uparrow$ 0.2)} &  \textbf{91.2 ($\uparrow$ 0.5)} &  \textbf{79.0 ($\uparrow$ 0.2)} & \textbf{67.7 ($\uparrow$ 0.4)} &  \textbf{75.9 ($\uparrow$ 0.4)} &  77.2 ($\downarrow$ 0.1) \\
\hline
   PoseFix \cite{PoseFix} & 32.475M & 8.992 &
   72.4 &  \textbf{91.0} &  79.9 & 69.1 & 78.5 & 77.8 \\
  \textbf{PoseFix-JRE}  & 32.476M & 8.993 &
   \textbf{72.6 ($\uparrow$ 0.2)} &  \textbf{91.0} &  \textbf{80.2 ($\uparrow$ 0.3)} & \textbf{69.4 ($\uparrow$ 0.3)} & \textbf{78.5} & \textbf{78.0 ($\uparrow$ 0.2)} \\
\hline
  UDP \cite{UDP} & 32.424M & 8.991 &
   71.2 &  \textbf{91.0} &  79.0 & 68.0 & 77.2 & 76.7 \\
  \textbf{UDP-JRE}  & 32.425M & 8.992 &
   \textbf{71.3 ($\uparrow$ 0.1)} &  90.9 ($\downarrow$ 0.1) &  \textbf{79.2 ($\uparrow$ 0.2)} & \textbf{68.1 ($\uparrow$ 0.1)} &  \textbf{77.3 ($\uparrow$ 0.1)} &  \textbf{76.8 ($\uparrow$ 0.1)} \\
\hline
  DarkPose \cite{DarkPose} & 32.424M & 8.991 &
   71.3 &  90.9 &  79.0 & 67.8 &  77.4 & 76.6 \\
  \textbf{DarkPose-JRE}  & 32.425M & 8.992 &
   \textbf{71.4 ($\uparrow$ 0.1)} &  \textbf{91.0 ($\uparrow$ 0.1)} &  \textbf{79.1 ($\uparrow$ 0.1)} & \textbf{68.0 ($\uparrow$ 0.2)} &  \textbf{77.5 ($\uparrow$ 0.1)} &  \textbf{76.8 ($\uparrow$ 0.2)} \\
\hline
\end{tabular}

\label{tab:coco}
\end{table*}

 \paragraph{\textbf{Quantitative analysis on the HiEve dataset.}} We evaluate the proposed RPSTN on the challenging HiEve dataset to demonstrate its advantages. Experimental results are listed in TABLE \ref{tab:hieve}. RPSTN outperforms DKD by 0.31 w-AP@avg and 0.32 AP@avg, and it surpasses HRNet by 0.63 w-AP@avg and 0.33 AP@avg. And other w-APs and APs of the RPSTN are improved over the DKD and HRNet, which proves the advantage of RPSTN to be the pose estimator. Besides, we also remove the JRE (RPSTN-w/o-JRE) to validate the effectiveness of the JRE. The performance of RPSTN is improved after integrating the JRE. The HiEve dataset is composed of crowded scenes. Human poses are frequently occluded. The correlation between joints is beneficial for locating the invisible joints, which indicates the effectiveness of the JRE.

\begin{table*}[htbp]
  \centering
  \footnotesize
  \setlength\tabcolsep{4.4pt}
  \caption{Experimental results on the HiEve dataset. The pretrained Faster-RCNN with ResNet-50 is used as the human detector. The values in brackets indicate the gap between the performance of the corresponding method and that of our RPSTN.}
  \begin{tabular}{lllllllll}
  \hline
  Methods & w-AP@avg &  w-AP@0.5 &  w-AP@0.75 & w-AP@0.9 & AP@avg & AP@0.5 & AP@0.75 & AP@0.9 \\
  \hline
   HRNet \cite{HRNet} &
   34.54 ($\downarrow$ 0.63) & 40.26 ($\downarrow$ 0.52) & 33.19 ($\downarrow$ 0.84) & 30.15 ($\downarrow$ 0.57) & 38.08 ($\downarrow$ 0.33) & 43.17 ($\downarrow$ 0.88) & 36.97 ($\downarrow$ 0.42) & 33.57 ($\downarrow$ 0.22) \\
   DKD \cite{9008554}&
   34.86 ($\downarrow$ 0.31) & 40.36 ($\downarrow$ 0.42) & 33.78 ($\downarrow$ 0.25) & 30.45 ($\downarrow$ 0.27) & 38.09 ($\downarrow$ 0.32) & 43.49 ($\downarrow$ 0.56) & 37.17 ($\downarrow$ 0.22) & 33.62 ($\downarrow$ 0.17) \\
   RPSTN-w/o-JRE &
   34.88 ($\downarrow$ 0.29) & 40.58 ($\downarrow$ 0.20) & 33.61 ($\downarrow$ 0.42) & 30.46 ($\downarrow$ 0.26) & 38.11 ($\downarrow$ 0.30) & 43.76 ($\downarrow$ 0.29) & 37.03 ($\downarrow$ 0.36) & 33.54 ($\downarrow$ 0.25) \\
  \textbf{RPSTN} &
   \textbf{35.17} & \textbf{40.78} & \textbf{34.03} & \textbf{30.72} & \textbf{38.41} & \textbf{44.05} & \textbf{37.39} & \textbf{33.79} \\
  \hline
  \end{tabular}
  \label{tab:hieve}
\end{table*}


 \paragraph{\textbf{Quantitative analysis on the COCO2017 dataset}} To verify the generality of JRE, we choose different human pose estimation models and evaluate the effectiveness of proposed JRE module on the challenging COCO2017 dataset. Experimental results on COCO2017 test-dev are shown in TABLE \ref{tab:coco}. The performance of the simple network (e.g., SBL) and the complex network (e.g., HRNet) is improved after incorporating the JRE module. It demonstrates that the JRE is appropriate for networks with different structures. We embed the JRE into S-ViPNAS that searched by NAS. The AP of S-ViPNAS is also improved, indicating that JRE is likewise suitable for the network searched by NAS. Furthermore, the proposed JRE improves the performance of the data preprocessing based method (e.g., UDP) and the heatmap post-processing-based method (e.g., DarkPose, PoseFix), proving that JRE is compatible with model-agnostic approaches. Moreover, we count parameters and GFLOPs for each backbone. It is clear that the proposed JRE adds a fixed amount of parameters (0.001M) and GFLOPs (0.001). PoseFix, UDP, and DarkPose are model-agnostic. Parameters and GFLOPs of the model only depend on the pose initializer. Because the SBL with ResNet-50 serves as the pose initializer in these methods, the parameters and GFLOPs of these methods are the same or similar to those of SBL. The official code of PoseFix is implemented by Tensorflow \cite{Tensorflow}. In contrast, other methods' official projects are based on Pytorch \cite{Pytorch}. This is the possible reason why there are slight differences in parameters and GFLOPs between the PoseFix and other model-agnostic approaches and SBL. But the additionally fixed parameters (0.001M) and GFLOPs (0.001) brought by JRE are consistent. In conclusion, experimental results suggest that the correlation between joints is beneficial for estimating poses. Furthermore, the proposed JRE module is a universal and plug-and-play module that is compatible with the mainstream pose estimation methods.

 \begin{figure}[htbp]
    \centering
   \includegraphics[scale=0.17]{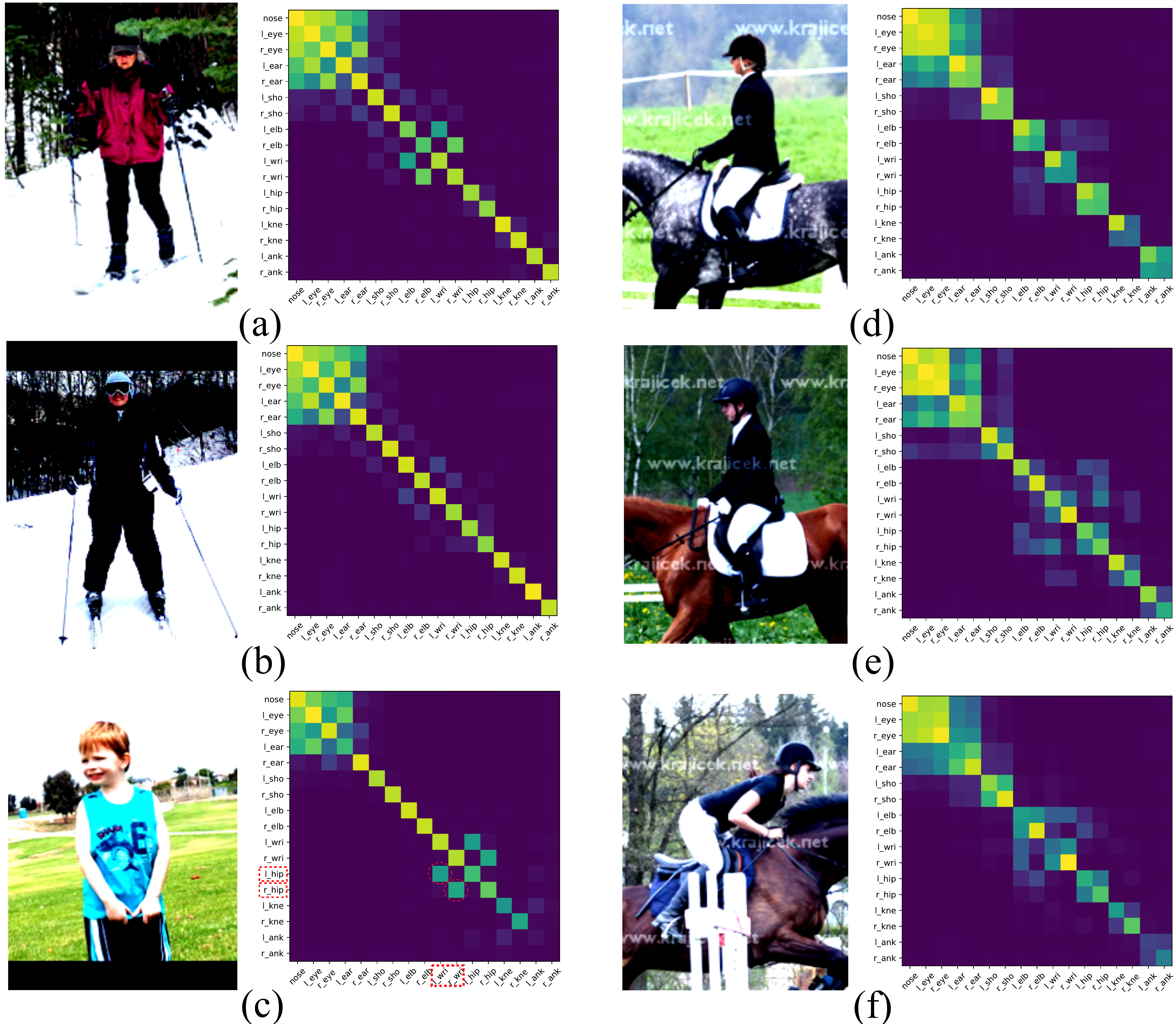}
   \caption{Visualization of the correlation matrix. Left: the correlation matrix for the action ``standing''. Right: the correlation matrix for the action ``horse-riding''.}
    \label{fig:cor_matrix}
 \end{figure}

 \begin{figure}[htbp]
    \centering
   \includegraphics[scale=0.23]{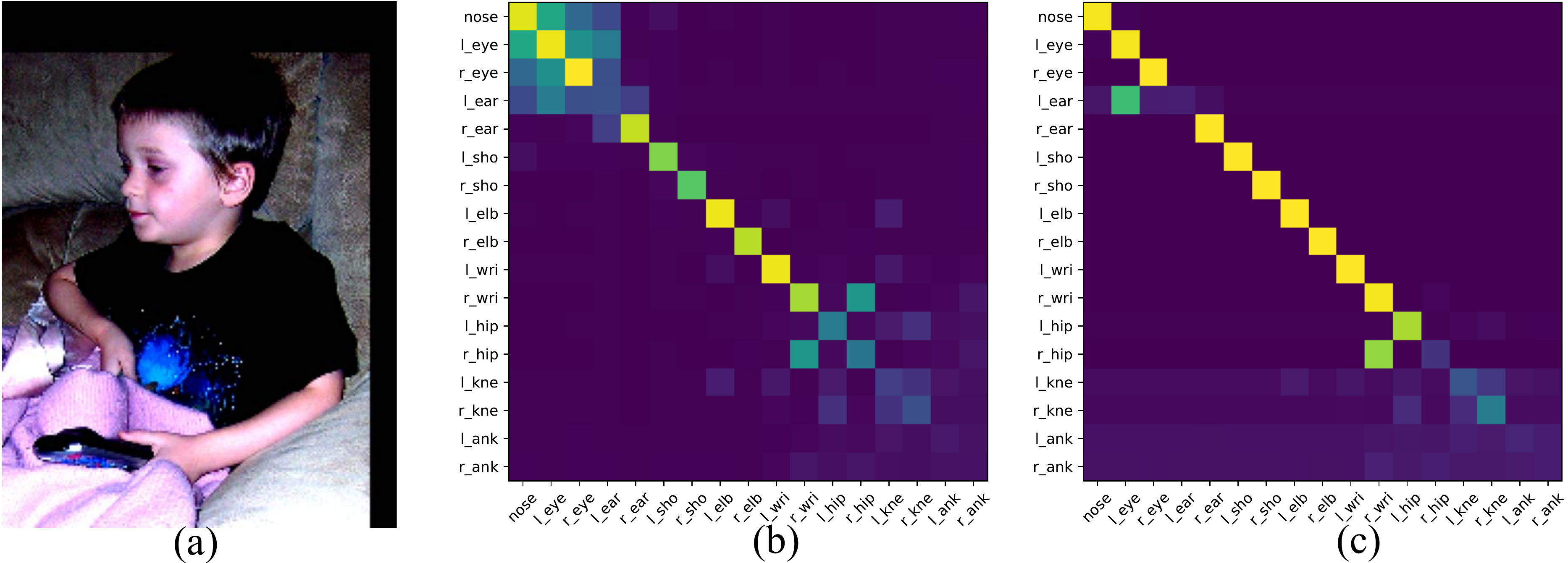}
   \caption{Visualization of the correlation matrix and the relational weight.From left to right, there are (a) the raw image, (b) correlation matrix, and (c) relational weight $W_{r}$, respectively.}
    \label{fig:wr}
 \end{figure}

 \begin{figure}[htbp]
    \centering
    \subfigure[The input (named as pseudo heatmaps) of the JRE module. Joints may exist in the highlighted area in the figure. The pseudo heatmap includes the spatial representation and the positional information of the joint. For the occluded joint, such as the ankle and the knee in the red dashed boxes, there may be inapparent highlighted areas in the pseudo heatmap.]{
        \centering
        \includegraphics[scale=0.34]{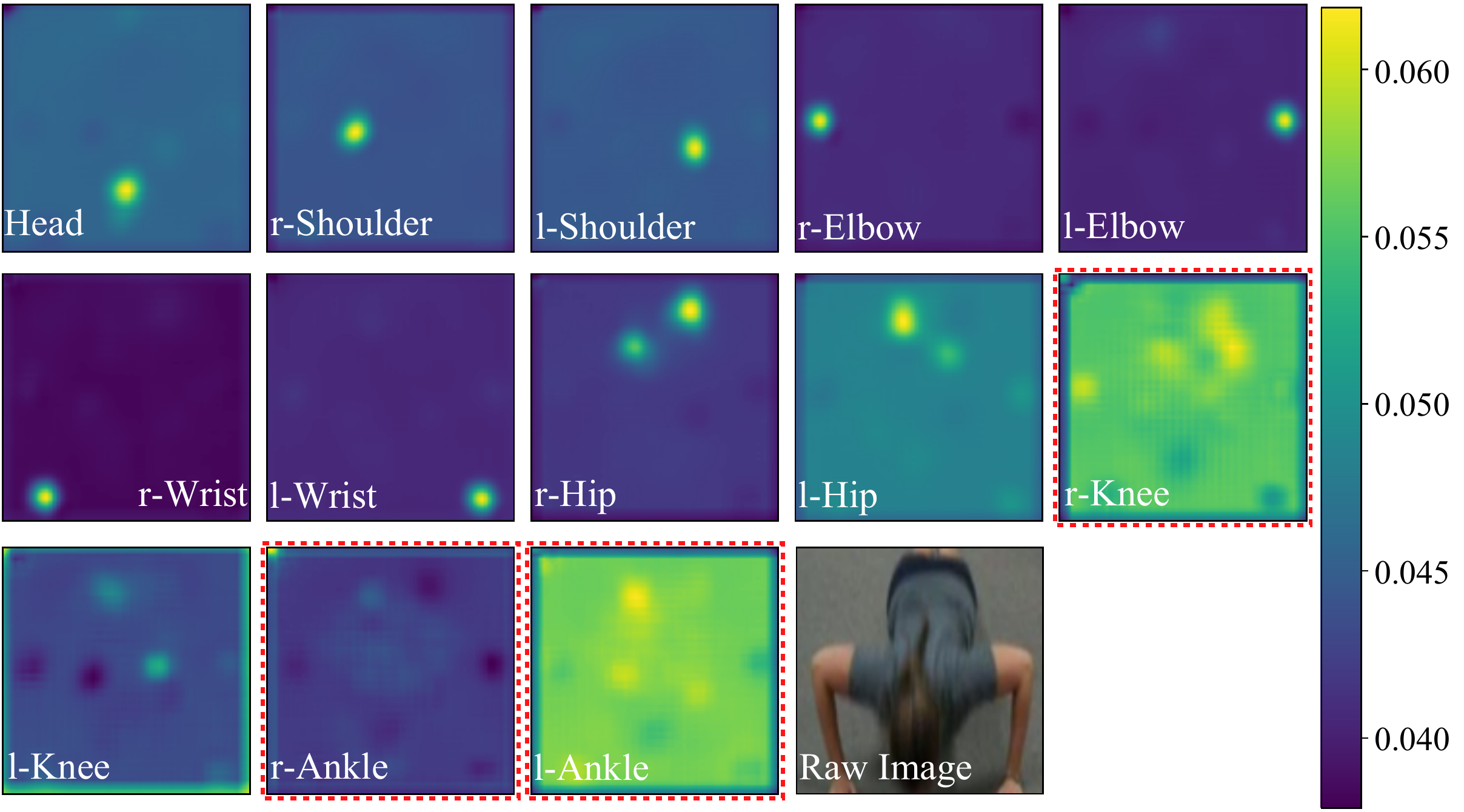}
        \label{fig:SKL:input}
    }

    \subfigure[The output of the JRE module. The JRE module models the structural information of poses by explicitly modeling the relationship between any two joints. The position of the occluded joint is inferred by modeling the occluded joint with other visible joints, such as the ankle and knee in the red dashed boxes.]{
        \centering
        \includegraphics[scale=0.34]{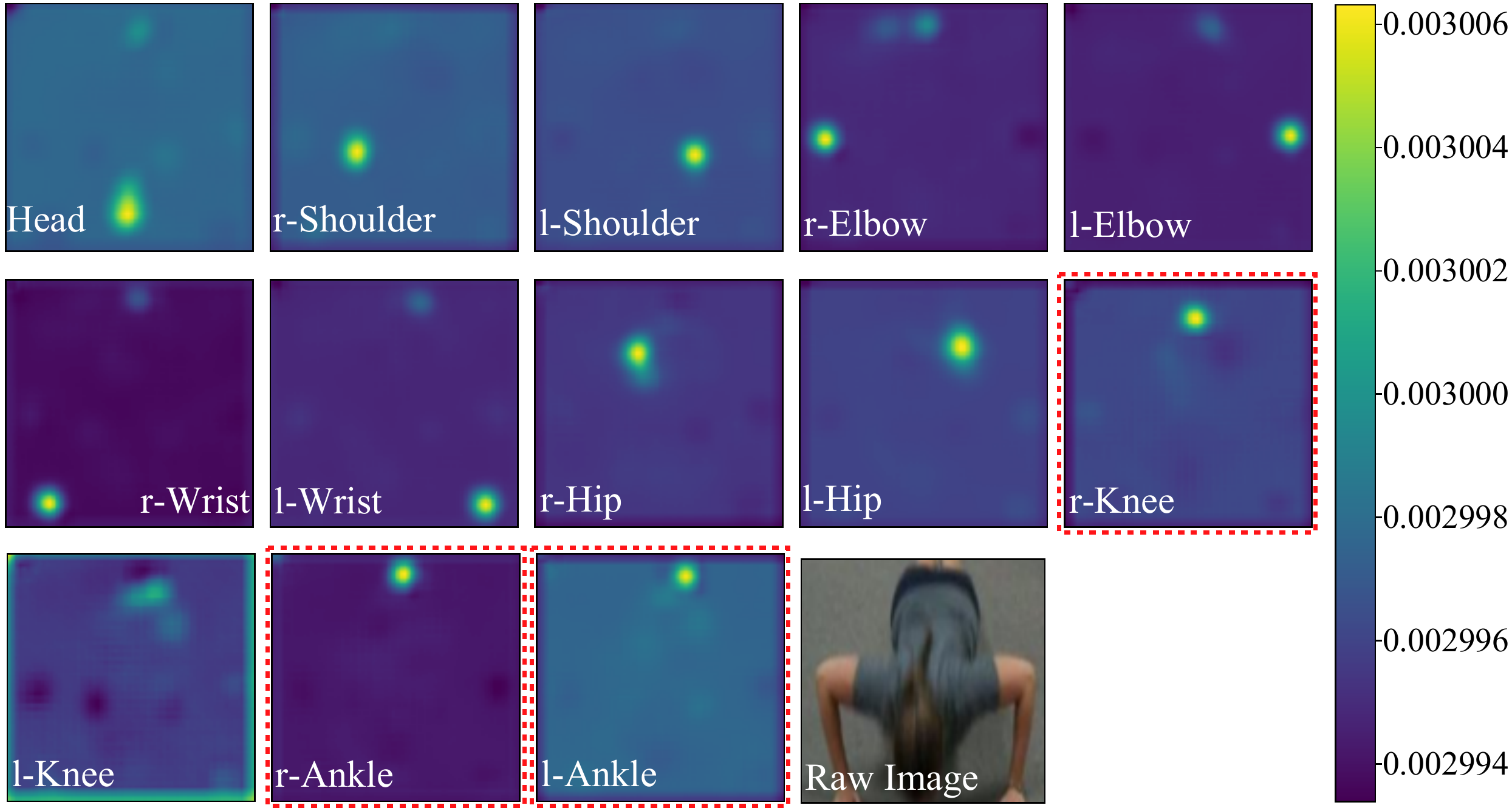}
        \label{fig:SKL:output}
    }
    \caption{The input and output of the JRE module. The 'r-' represents the right, and 'l-' represents the left.}
    \label{fig:SKL}
 \end{figure}

\begin{figure}[htbp]
    \centering
    \subfigure[Visualization on the Penn Action dataset. Following \cite{AID}, we randomly occlude some joints with the manual mask, and use the proposed method to estimate poses from occluded frames. Poses in (a) and (c) are the ground truth. Poses in (b) and (d) are estimated by the proposed RPSTN.]{
        \centering
        \includegraphics[scale=0.14]{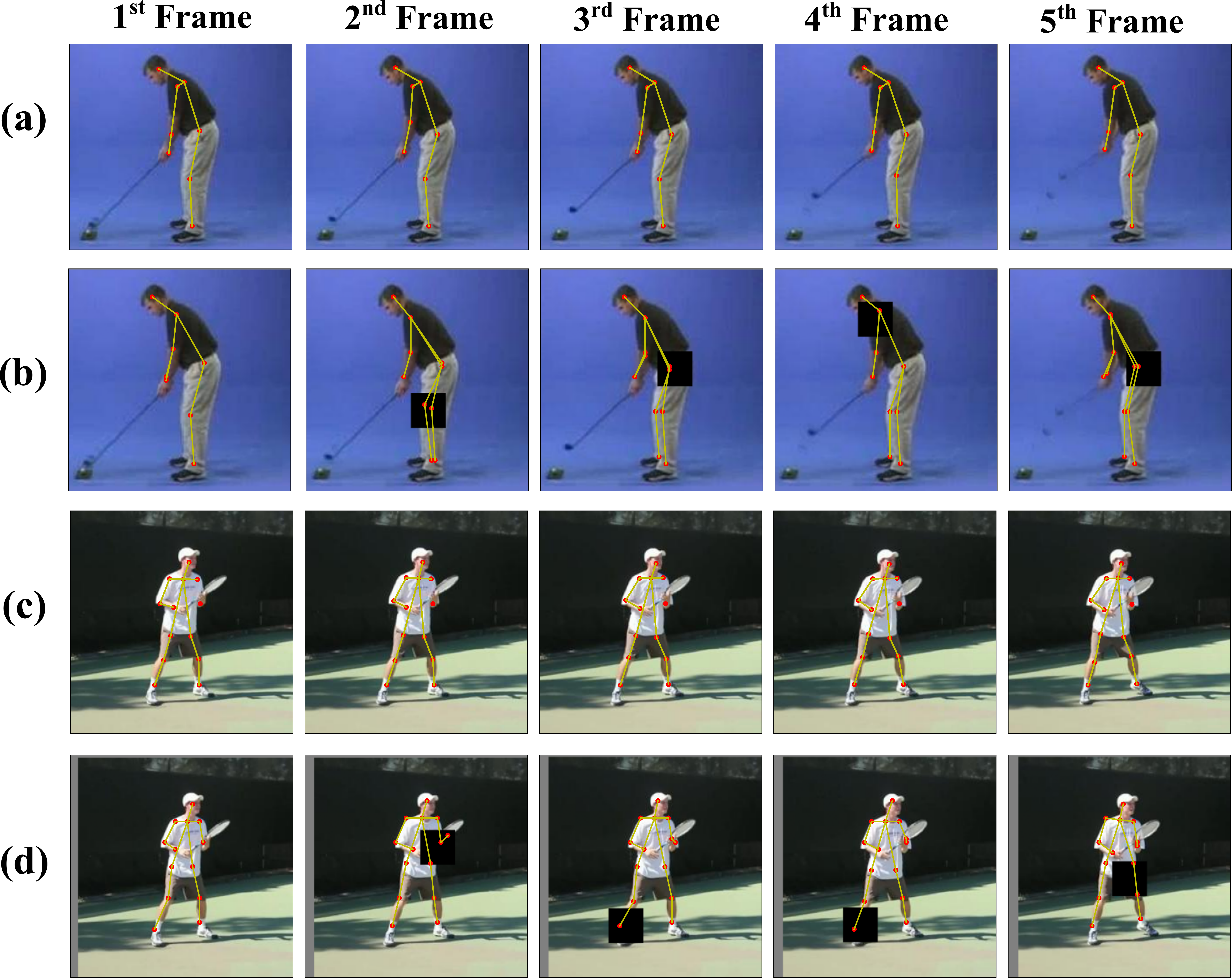}
        \label{fig:occlusion:penn}
    }

    \subfigure[Visualization on the COCO2017 dataset. Poses in row (a) are from SBL \cite{SimpleBaseline}. Poses in row (b) are from SBL-JRE. Red points represent invisible and unlabeled joints, blue points represent invisible but labeled joints, and green points represent visible and labeled joints. Keypoints in the red circles are that SBL-JRE locates more accurately than SBL.]{
        \centering
        \includegraphics[scale=0.14]{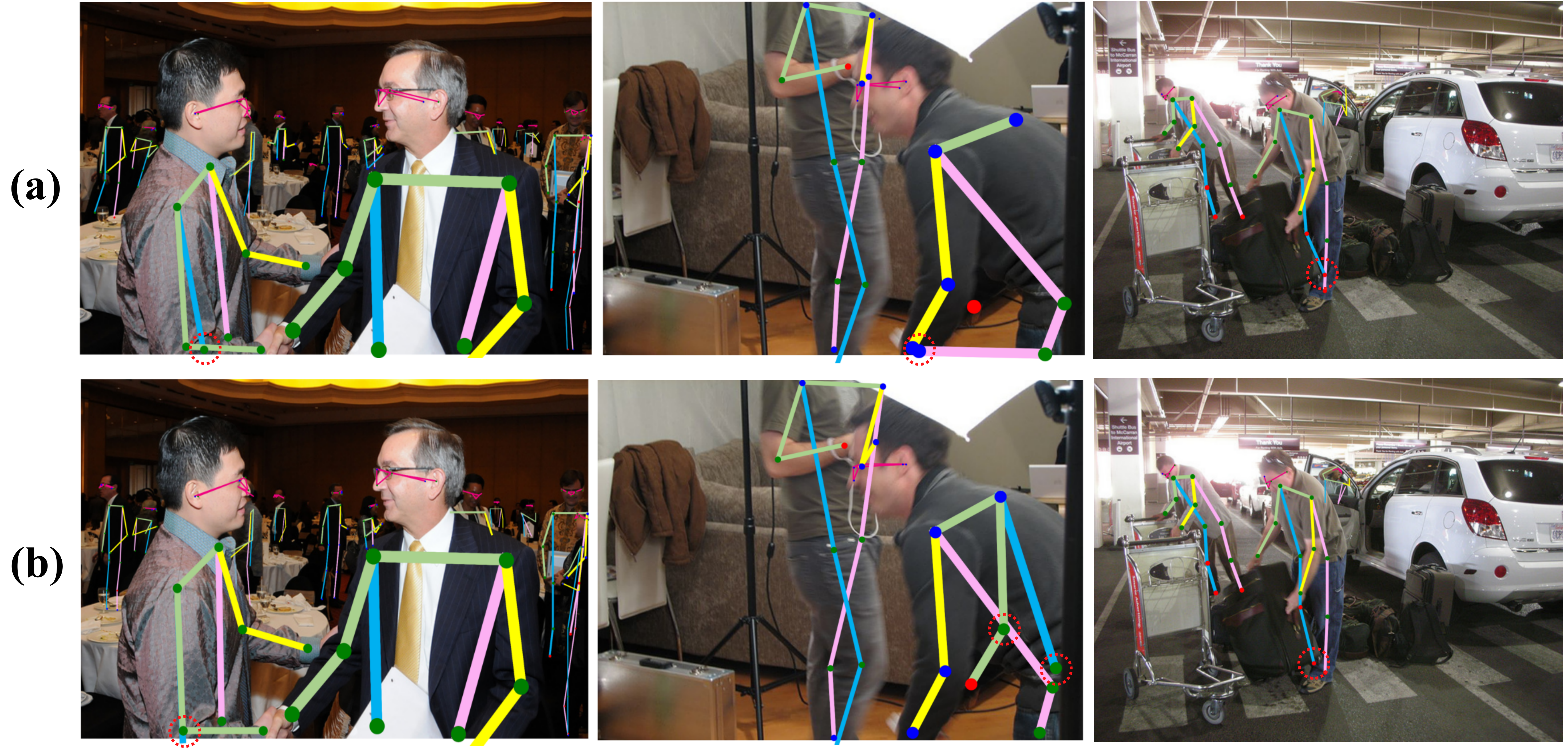}
        \label{fig:occlusion:coco}
    }
    \caption{Pose estimation for the occlusion situation.}
    \label{fig:occlusion}
 \end{figure}

\subsection{Qualitative Analyses}

 \paragraph{\textbf{Visualization of the correlation matrix}} In order to intuitively observe the correlation between joints modeled by the JRE, we visualize the correlation matrix as shown in Fig. \ref{fig:cor_matrix}. The correlation matrix models the correlation between any two joints by calculating the similarity of any two pseudo heatmaps. For the joints with high overall correlation, such as facial joints (the nose, eyes, and ears), the degree of correlation between them is high, as shown in Fig. \ref{fig:cor_matrix}, a highlighted region is formed between the facial joins. For those visible joints, their own correlations are high. For example, the standing person's shoulders, elbows are not occluded in Fig. \ref{fig:cor_matrix} left. There is only one highlighted point in the correlation matrix. For occluded joints, the correlation matrix highlights the joints related to the occluded joint. There are some highlighted rectangular areas in correlation matrix. As shown in Fig. \ref{fig:cor_matrix} (c), the boy's hips are occluded by wrists. The correlation matrix learns the correlation between hips and wrists. Half of the body is occluded in Fig. \ref{fig:cor_matrix} (d), (e), and (f). Highlighted areas are formed between the symmetrical joints.

 Besides, the correlation matrix is, to some extent, related to actions. For the same action, the correlation matrixes are usually similar, such as the correlation matrix in Fig. \ref{fig:cor_matrix}, (a), (b), and (c); (d), (e), and (f) . For different actions, the correlation matrixes are different to some degree. For example, there are some differences between the correlation matrix of the action ``standing" and that of the action ``horse-riding".

 \paragraph{\textbf{Visualization of the relational weight $W_{r}$}} Based on the correlation learned by the correlation matrix, $W_{r}$ models the correlation between the current joint and all joints, and can pay much attention to joints related to the current joint. For the visible joints that are easily located, the information of the visible joint itself is important. At the same time, their attention weight is high. As shown in Fig. \ref{fig:wr} (c), boy's shoulders, elbows, and wrists are visible. These joints has high attention weights. For the invisible joints that are difficult to be located, joints that are related to invisible joints are important. $W_{r}$ highlights the joints that are related to the occluded joint. As shown in Fig. \ref{fig:wr}, the boy's lower limbs are invisible. Other joints' position should be considered to assist in locating the knee and ankle.

 \paragraph{\textbf{Visualization of the input and output for the JRE module}} To better reveal the advantages of our JRE module, we visualize the input and output of the JRE module in Fig. \ref{fig:SKL}. As shown in Fig. \ref{fig:SKL:input}, the input of the JRE can be termed as the pseudo heatmap, which contains the spatial representation and positional information of the joint. For some occluded joints, such as knees and ankles in the red dashed boxes, joints' positional information is ambiguous. For these invisible joints, the structural prior of poses is used to help locate joints. The proposed JRE module can infer the position of the invisible joint by modeling the correlation between the occluded joint and other joints. As shown in Fig. \ref{fig:SKL:output}, by modeling the relationship between the 'r-knee' and other joints, the precise position of the 'r-knee' can be obtained.

 \paragraph{\textbf{Visualization of poses for the occlusion situation}} In order to show the robustness of the proposed RPSTN to the occlusion, we randomly occlude some joints following \cite{AID} and use RPSTN to estimate poses from occluded frames directly on the Penn Action dataset. Furthermore, to separately verify that the JRE module is beneficial for locating occluded joints, we use the original SBL and SBL \cite{SimpleBaseline} with JRE (SBL-JRE) to estimate poses directly from the COCO2017 dataset. We have visualized poses on the Penn Action dataset and the COCO2017 dataset, as shown in Fig. \ref{fig:occlusion}.

 For the Penn Action dataset, qualitative results are shown in Fig. \ref{fig:occlusion:penn}. We visualize the ground truth without masks (row 1) and the model's output with masks (row 2). The proposed RPSTN accurately locates the occluded joints. Furthermore, for some joints annotated as invisible, such as the right knee in the $2^{nd}$ frame and right ankle in the $3^{rd}$ frame, the RPSTN also infers a reasonable location for the joint. Experimental results prove that the proposed method can help deal with the occlusion situation. Moreover, the occluded right ankle is located in the $2^{nd}$ frame, and the positional information of the right ankle is transferred to the $4^{th}$ and $5^{th}$ frames.

 For the COCO2017 dataset, according to the visibility provided by the official annotations, we divide joints into three groups. As shown in Fig. \ref{fig:occlusion:coco}, red points represent invisible and unlabeled joints. Blue points represent invisible but labeled joints, and green points represent visible and labeled joints. Poses in Fig. \ref{fig:occlusion:coco}, row (a) are estimated by the SBL \cite{SimpleBaseline}. Poses in Fig. \ref{fig:occlusion:coco}, row (b) are estimated by the SBL-JRE. Keypoints in red circles are the joints that SBL-JRE locates more accurately than SBL. It can be seen that the JRE can help locate invisible joints by modeling the correlation between the invisible joint and its related joints. As shown in the $2^{nd}$ image in Fig. \ref{fig:occlusion:coco}, SBL does not locate the right elbow and right hip. However, after adding the JRE, the right elbow and right hip are reasonably located via the correlation between invisible joints and other joints. Moreover, the JRE can correct the unreasonable joints located by SBL, such as the right hip in the $1^{st}$ image in Fig. \ref{fig:occlusion:coco}, and right ankle in the $3^{rd}$ image in Fig. \ref{fig:occlusion:coco}. Experimental results show that JRE can improve the robustness of the model to the occlusion situation.

\begin{figure*}
   \centering
   \includegraphics[scale=0.18]{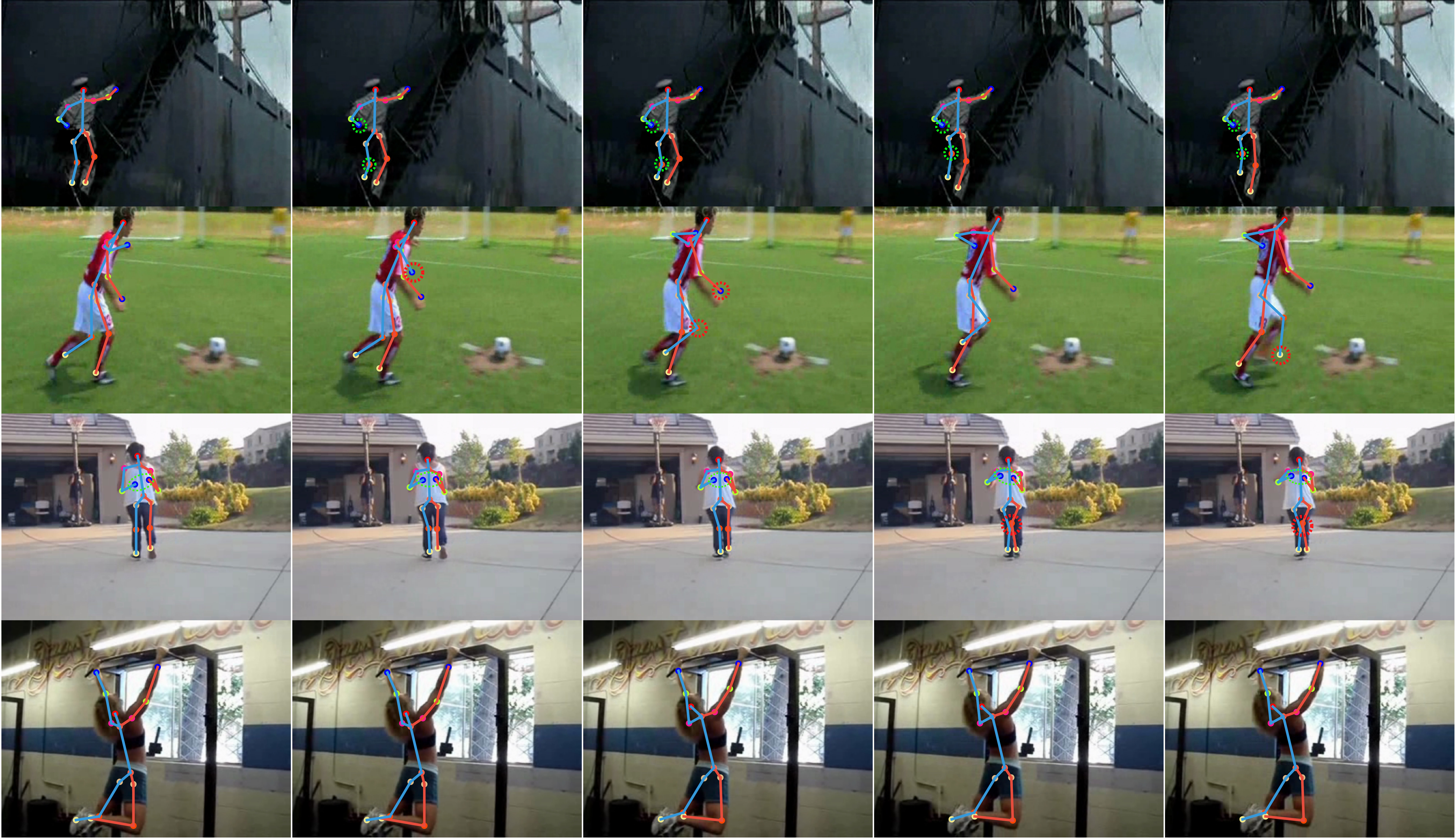}
   \caption{Visualization on the Sub-JHMDB dataset. The joints in green circles represent the occluded joints, and our model provides reasonable position for these joints. The joints in red circles represent that our model gives unreasonable position.}
   \label{fig:sub-jhmdb}
 \end{figure*}

 \paragraph{\textbf{Visualization of poses on the Sub-JHMDB dataset}} We perform visualization on the Sub-JHMDB dataset to further verify the effectiveness of the RPSTN, as shown in Fig. \ref{fig:sub-jhmdb}. In row 1 and row 3, joints in green circles are occluded. The proposed method transfers visible joints (the wrist and knee in row 1, the first frame) to the subsequent frames where wrist and knee are occluded. Positions of the occluded joints are obtained according to the visible information of joints in the first frame. Moreover, the proposed RPSTN can help locate the occluded joints according to the correlation between the occluded joints and other joints, as shown in Fig. \ref{fig:sub-jhmdb}, row 3. The body occludes two wrists, but our model provides a reasonable position. However, the performance of our method is limited by severe occlusion and the semantic information of joints. For example, due to the side view of the body, most joints are occluded in Fig. \ref{fig:sub-jhmdb}, row 2. It is difficult for the JRE module to learn the robust pose structural information, leading to poor performance for the joints in red circles.

\begin{figure*}[htbp]
    \centering
    \subfigure[PCK of each joint under different PCK thresholds $\gamma$ on the Penn Action dataset.]{
        \centering
        \includegraphics[scale=0.25]{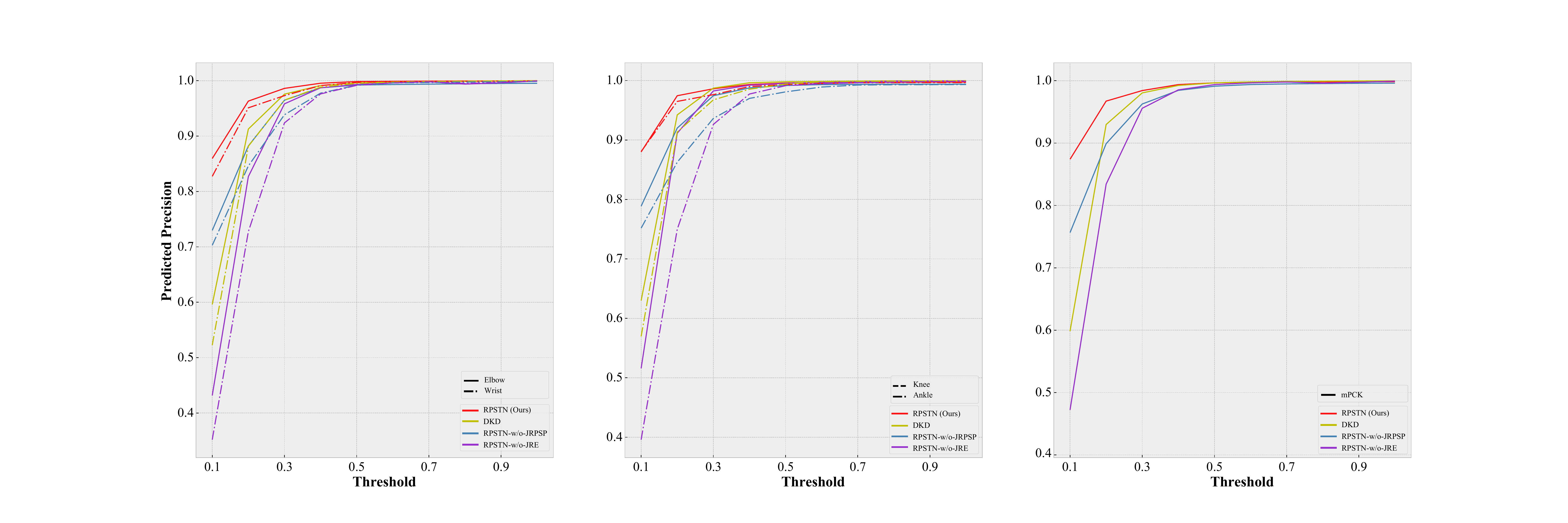}
        \label{fig:PCK:penn}
    }

    \subfigure[PCK of each joint under different PCK thresholds $\gamma$ on the Sub-JHMDB dataset.]{
        \centering
        \includegraphics[scale=0.25]{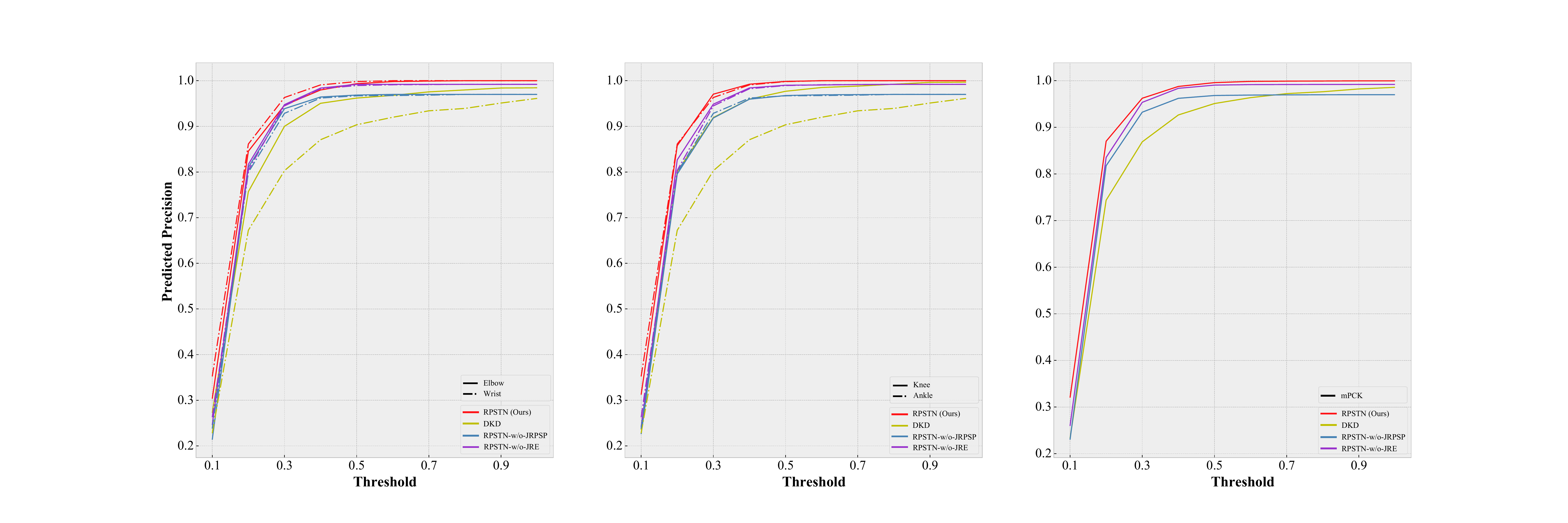}
        \label{fig:PCK:hmdb}
    }
    \caption{Visualization of the PCKs for joints under different thresholds $\gamma$. The $\gamma$ is initially set to 0.1 and increased to 1.0 with the step size of 0.1. From left to right, there are the PCKs of upper limb joins (left), the PCKs of lower limb joints (middle), and the mean PCK (right).Different colors represent different methods, i.e., DKD (yellow), RPSTN without JRPSP (blue), RPSTN without JRE (purple), and our RPSTN (red).}
    \label{fig:PCK}
 \end{figure*}

\paragraph{\textbf{Predicted precision under different PCK thresholds}} The PCK threshold $\gamma$ is usually set to 0.2. To evaluate the model's performance under different PCK thresholds, we increase the PCK threshold from  0.1 to 1.0. The predicted precision for each joint and the mean PCK are recorded in Fig. \ref{fig:PCK}. In order to intuitively observe the trend of the joint's PCK with the threshold, we retain the joints with a high flexibility, such as wrist, elbows, knees, and ankles. We split these joints and mPCK into three groups: the mPCK, the upper limb joints (elbows and wrists), and the lower limb joints (knees and ankles). Besides, we use the same type of lines for the same joint. For example, we use the dotted line to represent wrists. We use different colors to represent different methods including DKD (yellow), RPSTN without JRPSP (blue), RPSTN without JRE (purple), and our RPSTN (red). As shown in Fig. \ref{fig:PCK}, when the threshold $\gamma$ is set to 0.4, the PCK for each joint begins to saturate on two datasets. Our method shows obvious advantages on the challenging Sub-JHMDB dataset. The pose in the Sub-JHMDB is more complex than that in the Penn Action. The correlation between joints modeled by JRE is beneficial for locating joints.

\section{Conclusion}

 This paper proposes a general plug-and-play joint relation extractor (JRE) to explicitly model the correlation between joints for human pose estimation. We show that the JRE is compatible with the mainstream pose estimation backbones. With the guidance of the correlation learned by JRE, the model can locate the occluded joints reasonably. Furthermore, we improve the DKD that is designed for the JRE, i.e, joint relation guided pose semantics propagator (JRPSP), to model the temporal semantic continuity of videos by distilling and propagating relation-based semantic features of the whole pose between two adjacent frames. Based on JRE and JRPSP, we present a Relation-based Pose Semantics Transfer Network (RPSTN) to estimate human poses from videos. The proposed RPSTN also can infer the invisible joints of in the temporal by transferring the pose information from the non-occluded frame to the occluded frame. However, the proposed RPSTN has two disadvantages: first, the proposed method only models the temporal information between two continuous frames, limiting the model to extract long-term temporal features. Second, when the human body is severely occluded, such as half of the body is occluded or the person disappears completely, this method can not infer the occluded part well due to the lack of spatial information. We will improve the model to overcome these two disadvantages in the future.


\section*{Acknowledgment}

 This work was supported partly by the National Natural Science Foundation of China (Grant No. 62173045, 61673192), and partly by the Fundamental Research Funds for the Central Universities(Grant No. 2020XD-A04-2).

\ifCLASSOPTIONcaptionsoff
  \newpage
\fi
\bibliographystyle{IEEEtran}
\bibliography{IEEEexample}

\begin{IEEEbiography}[{\includegraphics[width=1in,height=1.25in,clip,keepaspectratio]{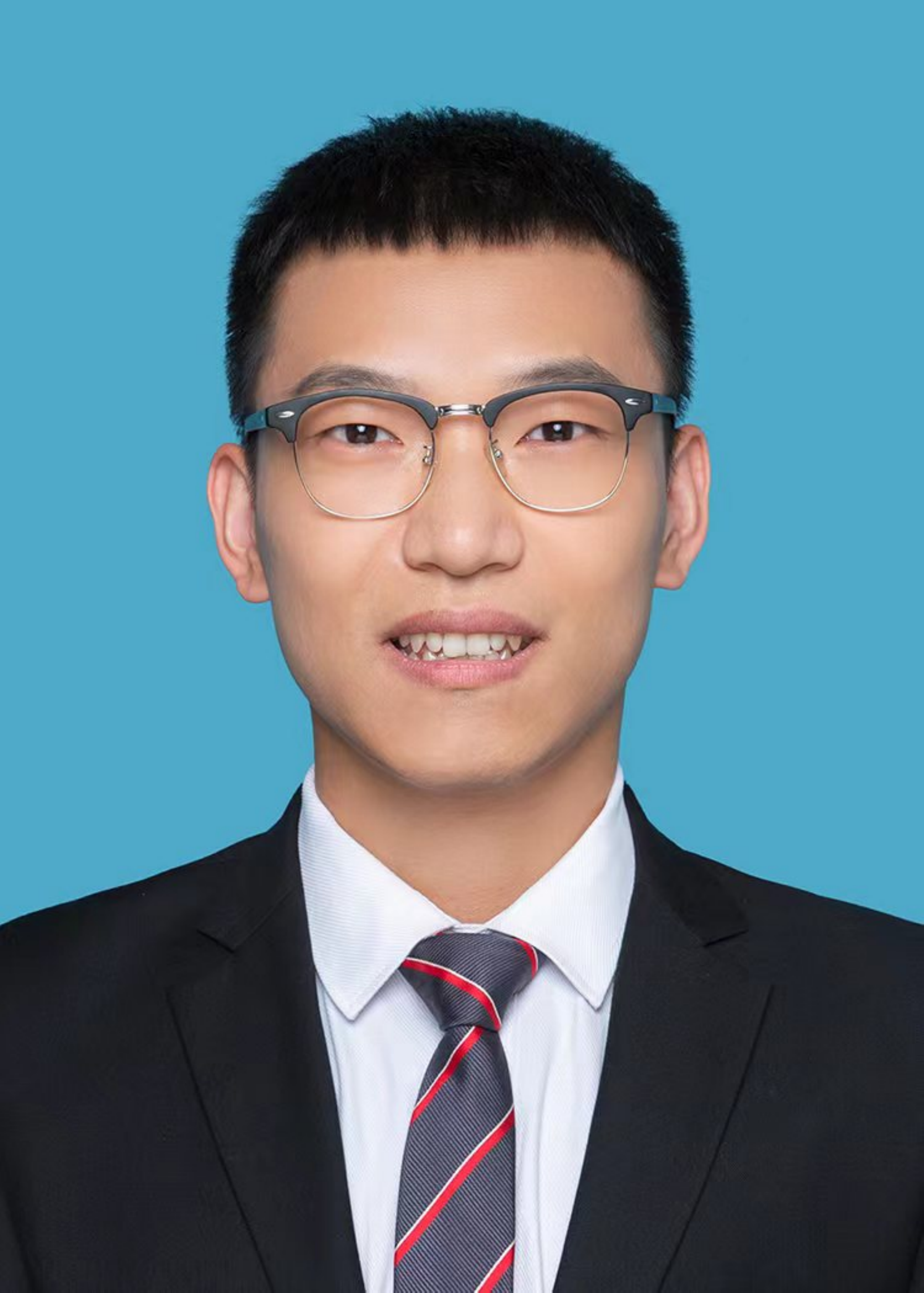}}]{Yonghao Dang}
Yonghao Dang received his Bachelor¡¯s degree in computer science and technology from the University of Jinan, Jinan, China, in 2018. He is currently a Ph.D. candidate with the School of Artificial Intelligence, Beijing University of Posts and Telecommunications, Beijing, China. His research interests include computer vision, machine learning and image processing, deep learning. Email: dyh2018@bupt.edu.cn.
\end{IEEEbiography}


\begin{IEEEbiography}[{\includegraphics[width=1in,height=1.25in,clip,keepaspectratio]{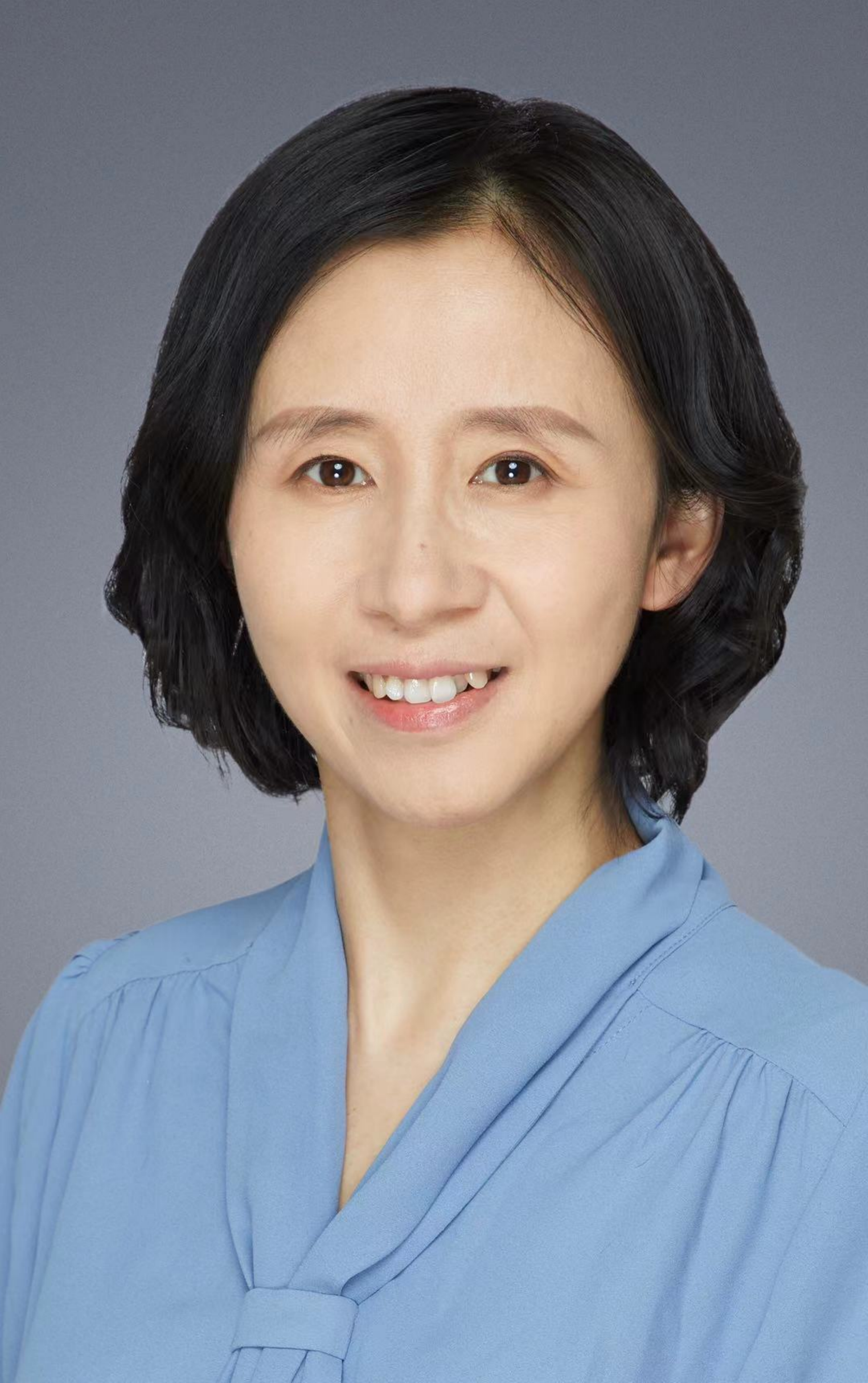}}]{Jianqin Yin}
 Jianqin Yin received the Ph.D. degree from Shandong University, Jinan, China, in 2013. She currently is a Professor with the School of Artificial Intelligence, Beijing University of Posts and Telecommunications, Beijing, China. Her research interests include service robot, pattern recognition, machine learning and image processing. Email: jqyin@bupt.edu.cn.
\end{IEEEbiography}

\begin{IEEEbiography}[{\includegraphics[width=1in,height=1.25in,clip,keepaspectratio]{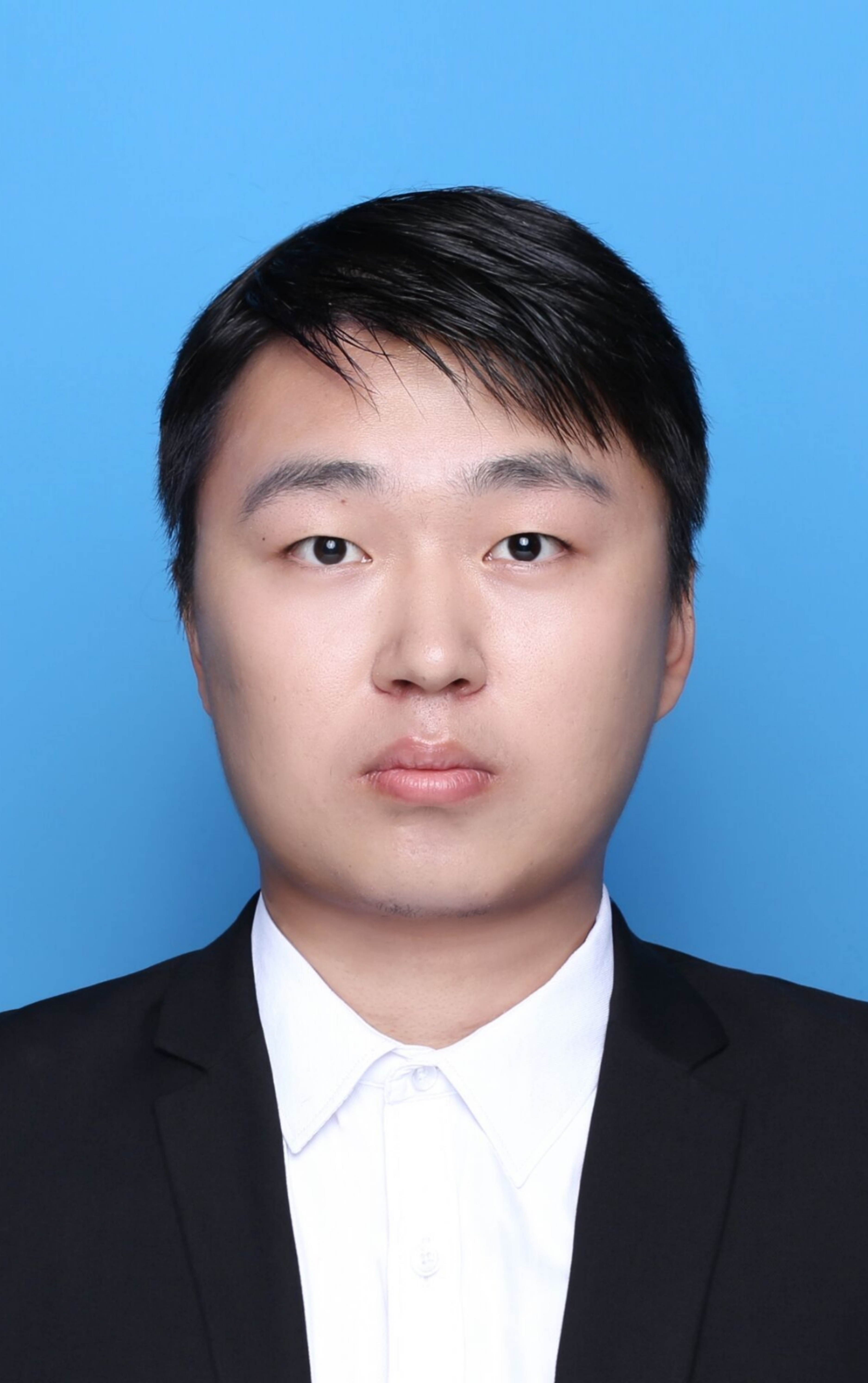}}]{Shaojie Zhang}
Shaojie Zhang is currently a Bachelors candidate with the School of Artificial Intelligence, Beijing University of Posts and Telecommunications, Beijing, China. His research interests include computer vision, deep learning. Email: zsj@bupt.edu.cn.
\end{IEEEbiography}




\end{document}